\documentclass[oneside, 11pt]{article}
\usepackage{iclr2024_conference,times}
\iclrfinalcopy

\usepackage[utf8]{inputenc} % allow utf-8 input
\usepackage[T1]{fontenc}    % use 8-bit T1 fonts
\usepackage{hyperref}       % hyperlinks
\usepackage{url}            % simple URL typesetting
\usepackage{booktabs, makecell}       % professional-quality tables
\usepackage[flushleft]{threeparttable}

\usepackage{amsfonts}       % blackboard math symbols
\usepackage{nicefrac}       % compact symbols for 1/2, etc.
\usepackage{microtype}      % microtypography
\usepackage{xcolor}         % colors

\usepackage{amssymb,amsmath, amsthm, amscd,ifthen, xfrac}
\usepackage{multirow}
\usepackage{graphicx}
\usepackage{tikz}
\usepackage{subcaption}
\usepackage{xcolor}
\usepackage{tabularx}
\usepackage{bbm}
\usepackage{comment}
\usepackage{framed}
\usepackage{transparent}
\usepackage{color}
\usepackage{dirtytalk}
\usepackage{framed}
\usepackage{cases}
\usepackage{cancel}
\usepackage{mdframed}
\usepackage{empheq}

\usepackage{enumitem}

\usepackage{natbib}
\definecolor{darkgreen}{RGB}{0, 150, 0}

\def\T{\top}
\def\E{\mathbf{E}}
\def\R{\mathbb{R}}
\def\Tr{\mathrm{Tr}}
\def\Var{\mathrm{Var}}

\def\uv{\mathbf{u}}
\def\gv{\mathbf{g}}

\newtheorem{lemma}{Lemma}
\newtheorem{remark}{Remark}

\newtheorem{corollary}{Corollary}
\newtheorem{theorem}{Theorem}

\title{Revisiting inverse Hessian vector products for calculating influence functions}

\author{%
  Yegor Klochkov \\
  ByteDance Research\\
  \texttt{yegor.klochkov@bytedance.com} \\
  \And
  Yang Liu \\
  CSE, UC Santa Cruz\\
  \texttt{yangliu@ucsc.edu} \\
}

\begin{document}

\maketitle

\begin{abstract}
    %Influence functions are a popular tool for attributing model's output to training data. The traditional approach relies on the calculation of the inverse Hessian-vector products (iHVP), but the classical solver ``Linear time Stochastic Second-order Algorithm'' (LiSSA, \cite{agarwal2017second}) is often deemed impractical for large models due to expensive computation and hyperparameter tuning. We show that the three hyperparameters --- the scaling factor, the batch size, and the number of steps, can be chosen depending on the spectral properties of the Hessian, particularly, its trace and the largest eigenvalue. By evaluating with random sketching \citep{swartworth2023optimal}, we find that the batch size has to be sufficiently large for  LiSSA to converge, however, for all of the models that we consider the requirement is mild. We confirm our findings  empirically by comparing to Proximal Bregman Retraining Functions (PBRF, \cite{bae2022if}). Finally, we discuss what role does the inverse Hessian play in calculating the influence.
    Influence functions are a popular tool for attributing a model's output to training data. The traditional approach relies on the calculation of inverse Hessian-vector products (iHVP), but the classical solver ``Linear time Stochastic Second-order Algorithm'' (LiSSA, \cite{agarwal2017second}) is often deemed impractical for large models due to expensive computation and hyperparameter tuning. We show that the three hyperparameters --- the scaling factor, the batch size, and the number of steps --- can be chosen depending on the spectral properties of the Hessian, particularly its trace and largest eigenvalue. By evaluating with random sketching \citep{swartworth2023optimal}, we find that the batch size has to be sufficiently large for LiSSA to converge; however, for all of the models we consider, the requirement is mild. We confirm our findings empirically by comparing to Proximal Bregman Retraining Functions (PBRF, \cite{bae2022if}). Finally, we discuss what role the inverse Hessian plays in calculating the influence.
\end{abstract}

\section{Introduction}

Deep neural networks have seen many impressive results in the past years, but researchers and practitioners have little understanding what happens inside the models and how they learn to predict. While there are exist various methods to interpret the internal computations in an understandable to a human way, influence functions attempt to explain model behaviour by  attributing model predictions (or generations) to particular examples in the training data. 

\cite{koh2017understanding} introduce Hessian-based influence functions in order to approximate the effect of removal of one training point from the training set, which we refer to as leave-one-out retraining. The formula for influence calculation is derived from the second-order Taylor approximation of the loss, hence the Hessian and the gradient of the training point are sufficient for calculation. \cite{koh2017understanding} demonstrate various applications of influence functions such as explaining of model outputs through data attribution, repairing mislabeled data, and backdoor attacks.

\cite{basu2020influence} criticize influence functions for poor approximation of leave-one-out retraining as depth and width of neural networks increase. As a solution, \cite{bae2022if} propose two fixes: replacing the Hessian (that possibly has negative eigenvalues) with well-behaved Gauss-Newton Hessian \citep{martens2020new}, and replacing the leave-one-out retraining (which itself is not a well-defined objective) with Proximal Bregman Retraining Functions (PBRF). They demonstrate that the latter do not suffer from the randomness introduced by model initialization and data sampling, and they argue can serve as gold standard when evaluating influence function approximation methods. In this paper, we focus on this particular formulation of influence functions, where the Hessian is replaced with Gauss-Newton Hessian, and the PBRF serves as the ground truth in validation.

The calculation of influence functions, as introduced by \cite{koh2017understanding}, requires approximation of inverse Hessian-vector products. Given the dimension of modern deep models and the size of the training dataset, it can be a hard problem. As an alternative to traditional conjugate gradient method, \cite{koh2017understanding} proposed a stochastic iterative approach called ``Linear time Stochastic Second-Order Algorithm'', or in short LiSSA \citep{agarwal2017second}. This algorithm requires calculating a sampled Hessian-vector product at each iteration, where in the formulation of \cite{koh2017understanding}, the batch size per sample can be as little as just one training point. There are two additional hyperparameters involved --- the scaling factor and the number of steps in LiSSA, with little direction of how to choose them in practice. \cite{basu2020influence} criticize LiSSA, suggesting that it lacks convergence for deep networks with large number of parameters. In particular, the method deemed impractical due to the need for expensive hyperparameter search. In this paper, we carefully analyze the convergence of LiSSA and find that the choice of all three hyperparameters, including the batch size, depends on the properties of the Gauss-Newton Hessian, namely, its trace and largest eigenvalue. Since the size of the Hessian is very large (number of parameters to square), we evaluate these two statistics with \emph{random sketching} \citep{swartworth2023optimal}, which only requires estimation of Hessian-vector products in the process. We report these statistics and the corresponding requirements for some open sourced vision and language models. We also note that, contrary to a common belief, we find that the batch size has to be sufficiently large for the algorithm to converge. However, this requirement is mild, and for language models it is particularly redundant.

Some attempts to avoid calculating inverse Hessian vector products have been made in the literature. \cite{schioppa2022scaling} suggest to truncate the spectrum of the Hessian, see also \cite{fisher2023influence,  grosse2023studying}. 
When it comes to language models, most of the recent literature is using  gradient-based influence functions \citep{xia2024less, he2024s, chhabra2024outlier}. These are typically focused on the finetunning stage, and often this choice is motivated by simpler and faster implementation.
In Section~\ref{sec:hess_role}, we highlight the difference between the results of these two ways of calculating influence functions.

Almost exclusively, \cite{grosse2023studying} calculate Hessian-based influence of pretraining data for LLMs. Their   analysis is restricted to MLPs of the transformer and they impose a block-wise structure onto the Hessian. Although we do not advocate against such structural assumptions, our work suggests that running the plain and model-agnostic LiSSA can be  feasible, given that we avoid the hyperparameter search.
In our implementation, we follow the Hessian-free approach of \cite{martens2010deep} using finite differences, where an in-batch GNH-vector product is calculated with three forward propagations and one backward propagation. For example, our code allows to run LiSSA for 7B language models on a 4$\times$A100 GPU node.
We share the code here \url{https://github.com/yklochkov-bytedance/gnhtools}.%, it can run e.g. 7B models on 4 $\times$ 80GB GPUs. 

\def\xv{\mathbf{x}}
\def\sf{\mathrm{sf}}
\section{Background and notation}

Influence functions \citep{koh2017understanding} are calculated under the assumption that the optimized parameter $\theta$ of the model delivers minimum to the training loss,
\begin{equation}\label{loss_min}
    \theta^{*} = \arg\min_{\theta} \frac{1}{|\mathcal{D}_{tr}|} \sum_{(\mathbf{x}, y) \in \mathcal{D}_{tr}} \ell(\mathbf{x}, y; \theta),
\end{equation}
where for classification tasks, $ (\mathbf{x}, y) $ are input and label pair, and for language modeling tasks, consists of context and next word token. That is, given a sequence $ s = (s_1, \dots, s_l) $, the dataset $\mathcal{D}_{tr}$ consists of pairs $ \mathbf{x} = (s_1, \dots, s_{t-1})$ and $y = s_{t} $. Let us fix a point $ (\xv_m, y_m) \in \mathcal{D}_{tr} $, and for a small perturbation weight $\epsilon > 0$ consider
\begin{equation}\label{loo_formula}
    \theta^{*}(\epsilon) = \arg\min_{\theta} \frac{1}{|\mathcal{D}_{tr}|} \sum_{(\mathbf{x}, y) \in \mathcal{D}_{tr}} \ell(\mathbf{x}, y; \theta) + \epsilon \, \ell(\mathbf{x}_m, y_m; \theta).
\end{equation}
Then, the \emph{influence of a training point} $ (\xv_m, y_m) $ on the parameter is denoted as
\begin{align}
    \mathcal{I}((\xv_m, y_m)) &= \frac{d \theta^*(\epsilon)}{d \epsilon} \Big\vert_{\epsilon = 0} \nonumber \\
    &= - H^{-1} \nabla \ell(\xv_m, y_m; \theta^*), \label{influence_ihvp_formula}
\end{align}
where $H$ denotes the \emph{population Hessian}, that is
\[
    H = \frac{1}{|\mathcal{D}_{tr}|} \sum_{(\mathbf{x}, y) \in \mathcal{D}_{tr}} \nabla^{2} \ell(\mathbf{x}, y; \theta^*) \,.
\]

Furthermore, suppose we have a set of predictions $ z_{test} = (\xv_{test}, \hat{y}_{test})$, and let $ f((\xv, y), \theta) = \log p(y| \xv; \theta) $ are the log probability according to the trained model. Then, the influence of a \emph{training point $ z_{train} = (\xv_m, y_m)$ on a prediction $z_{test}$} is denoted as
\[
    \mathcal{I}(z_{train}, z_{test}) = - \nabla f(z_{test}, \theta)^{\T} H^{-1} \nabla \ell(\xv_m, y_m; \theta^*)
\]
For language models, we calculate the influence for completion following \cite{grosse2023studying}. Let $ s = (s_1, \dots, s_p) $  be a prompt and $ \hat{s} = (\hat{s}_1, \dots, \hat{s}_c) $ be a completion. Then, we calculate the influence for average log-probability of predicted tokens
\[
    f(s, \hat{s}; \theta) = \frac{1}{c} \sum_{j = 1}^{c} \log p(\hat{s}_j | s_1\dots s_p \hat{s}_1 \dots \hat{s}_{j-1} ; \theta)\,.
\]

\paragraph{PBRF as ground truth.} The inverse problem $H^{-1} \nabla \ell(z_{train}; \theta)$ can be difficult to perform due to degenerate eigenvalues of $ H $. \cite{koh2017understanding} propose to use a damping parameter $\lambda > 0$ and instead invert a regularized matrix $(H + \lambda I)^{-1}$.
However, such matrix can still be degenerate due to possibly negative eigenvalues of the Hessian of a non-convex loss, which are indeed observed in practice and are not necessarily small \citep{sagun2017empirical, schioppa2024gradient}.
Motivated by natural descent methods, \cite{bae2022if} propose to replace it with \emph{Gauss-Newton Hessian} (GNH), which is denoted as follows. Suppose that the loss has the form
\[
    \ell((\xv, y); \theta) = \ell(h(\xv; \theta), y), \qquad \ell(h, y) = - \log ( \sf(h)_{y}),
\]
where $ h(\xv; \theta) \in \R^{K} $ is the logit function and $ \sf(h)_j = \exp(h_j) / \left( \sum_{j = 1}^{K} \exp(h_j) \right) $ is the standard softmax function. Then, the GNH has the form
\begin{equation}\label{gnh_formula}
    H = \frac{1}{|\mathcal{D}_{tr}|} \sum_{(\mathbf{x}, y) \in \mathcal{D}_{tr}} [J_{\theta} h(\xv; \theta)] \nabla_{h}^{2} \ell(h(\mathbf{x}; \theta), y) [J_{\theta} h(\xv; \theta)]^{\T}
\end{equation}
For the Cross-Entropy loss, we have the identity $ \nabla_{h}^{2} \ell(h(\mathbf{x}; \theta), y) = \mathrm{Diag}(\sf(h)) - \sf(h) \sf(h)^{\T} $.

Furthermore, \cite{bae2022if} show that if the Hessian is replaced with the Gauss-Newton Hessian, the influence functions \eqref{influence_ihvp_formula} approximate a different retraining functions called Proximal Bregman Retraining Functions (PBRF). These correspond to retraining of the Proximal Bregman Objective (PBO) on training point $(\xv_m, y_m)$ reads as follows,
\begin{equation}
    \theta_{PBRF}(\epsilon) = \arg\min_{\theta} \frac{1}{|\mathcal{D}_{tr}|} \sum_{(\xv, y) \in \mathcal{D}_{tr}} D(h(\xv; \theta), h(\xv; \theta^{\star}), y) + \epsilon \ell((\xv_m, y_m); \theta) + \frac{\lambda}{2} \| \theta - \theta^{\star}\|^{2}, \label{pbrf_equation}
\end{equation}
where $D(h, h', y) = \ell(h, y) - \ell(h', y) - (h - h')^{\T} \nabla_{h} \ell(h', y)$ is the Bregman divergence. Comparing PBO with the objective in \eqref{loo_formula}, we see that the proximity penalty $\frac{\lambda}{2} \| \theta - \theta^{\star}\|^{2}$ takes into account the damping parameter, while replacing the loss with the Bregman divergence accounts for potential lack of convergence, i.e. we no longer need to assume that the training of the original parameter converges to global minimum of the loss as in \eqref{loss_min}. In addition, \cite{bae2022if} find that PBRFs are a more reliable objective compared to traditional retraining, which is known to produce different outputs \citep{basu2020influence}.

\cite{bae2022if} argue that PBRF is suitable ground truth objective for validation of influence function estimation algorithms. For instance, they use it for empirical confirmation of their ad-hoc algorithm in \cite{grosse2023studying}. Following them, we refer to PBRF as a ground truth influence in order to confirm our findings empirically.

\paragraph{Iterative inverse Hessian-vector products.} 

For calculating these inverse Hessian-vector products (iHVP) of form $\uv = (H + \lambda)^{-1}\gv$, \cite{koh2017understanding} propose to use a variant of Linear time Stochastic Second-Order Algorithm (LiSSA, \cite{agarwal2017second}), that consist of the iterations
\begin{equation}\label{lissa_update}
    \uv^{t} = \gv + (I - \eta (\tilde{H}^{t} + \lambda)) \uv^{t-1}, \qquad t = 1, \dots, T
\end{equation}
where $ \tilde{H}^{t} $ is an in-batch estimate of $H$. Ideally, the scaling parameter $\eta > 0$  needs to be chosen to ensure that $\eta H$ is a contraction, however, it requires knowning the larges eigenvalue of $H$. To this day, LiSSA is often discarded due to it's hyperparameters, which are not trivial to tune when one does not have a clear objective \cite{basu2020influence}. In particular, the three hyperparameters --- $\eta$, $T$, and the batch size, are often chosen without any directive, and the resulting estimate is deemed unreliable \cite{basu2020influence}.

The LiSSA updates \eqref{lissa_update} are equivalent to stochastic gradient descent  (SGD) with step size $\eta$ for the quadratic objective $
\min_{\uv} \frac{1}{2} \uv^{\T} (H + \lambda) \uv - \uv^{\T} \gv $ \citep{fisher2023influence}. In theory, mini-batch SGD is known to work at least as well as full gradient, even in terms of number of updates \citep{harvey2019tight}. However, larger batch sizes are often preferred by practitioners. The problem stems from theoretical results relying on a notion of uniform smoothness, which is not observed in practice, where an in-batch function is usually not as smooth as the average over the whole dataset \citep{tang2020practicality}. %\footnote{For example, the function $\| x \|^{2} = \frac{1}{d} \sum_{j = 1}^{d} x_j^2, x \in \R^{d}$ has smoothness $1/d$, while the smoothness of a single term $f_j(x) = x_j^2$ is $1$.}. In addressing this quesion,  find that mini-batch SGD can still be useful in the situations where the Hessian's eigenspectrum is rapidly decaying. We therefore focus on confirming the fast-decaying spectrum of the Hessian empirically.
Furthermore, optimal choice of step size $\eta$ and the number of steps $T$ depend on the largest eigenvalue of the Hessian $\lambda_{\max}(H)$. Rather than conducting hyperparameter tuning, we suggest to evaluate the largest eigenvalue $\lambda_{\max}(H)$ directly, which allows us to run the LiSSA just once.% \textcolor{red}{In Section~\ref{},} we conduct empirical analysis of the convergence of LiSSA and confirm that indeed for mini-batch implementation the fluctuations are negligible if the step size is chosen correctly.

\paragraph{Hessian-vector products.} The updates \eqref{lissa_update} involve calculation of in-batch Hessian-vector products $\tilde{H}_t \uv$. Expanding the expression for GNH \eqref{gnh_formula}, we observe that for a batch of data $B = \{ (\xv, y)\}$, we have
\begin{equation}\label{formula-tilde-H-dot-u}
    \tilde{H}_t \uv = \frac{1}{|B|} \sum_{(\mathbf{x}, y) \in B} [J_{\theta} h(\xv; \theta)] \nabla_{h}^{2} \ell(h(\mathbf{x}; \theta), y) [J_{\theta} h(\xv; \theta)]^{\T} \uv
\end{equation}
Here $ [J_{\theta} h(\xv; \theta)]^{\T} \uv $ is a directional derivative of vector-function $ h(\xv; \theta)$. Calculating the directional derivatives per each example in the batch precisely may be prohibitively expensive. Instead, we suggest to approximate it by finite differences,
\[
    [J_{\theta} h(\xv; \theta)]^{\T} \uv = \frac{d}{d \delta} h(\xv; \theta + \delta \uv) \Big\vert_{\delta = 0} \approx \frac{h(\xv; \theta + \delta \uv) - h(\xv; \theta - \delta \uv)}{2 \delta},
\]
where $\delta$ is a small value, which we fix to $\delta = 0.01$ in our experiments. Then, we can approximate the in-batch GNH-vector prodcut by using three forward propagations and one backward propagation:
\[
    \tilde{H}_t \uv \approx \nabla_{\theta} \frac{1}{|B|} \sum_{(\mathbf{x}, y) \in B} h(\xv; \theta)^{\T} S_{h} \left\{\frac{h(\xv; \dot{\theta} + \delta \uv) - h(\xv; \dot{\theta} - \delta \uv)}{2 \delta}\right\},
\]
where $\dot{\theta}$ indicates that we do not calculate the derivative through this parameter, and $S_h = \mathrm{Diag}(\sf({h})) - \sf({h})\sf({h})^{\T}$ is also fixed. Thus, we simply backpropogate through a weighted sum of logits in the batch $h(\xv; \theta)$, with weights depending on the matrices $S_h$ and finite differences $ (h(\xv; \dot{\theta} + \delta \uv) - h(\xv; \dot{\theta} - \delta \uv))/ (2 \delta) $. The latter two can be calculated in a gradient free manner with three forward propagations. 
We note that incorporating finite differences for Hessian-vector products calculation has previously been done in \cite{martens2010deep, martens2011learning}.

%\begin{remark}
%    \textcolor{red}{Remark on code and requirements and limitations. Notice that we don't have to flatten. Update when code is revisited. }
%\end{remark}

\section{Convergence of LiSSA and choice of hyperparameters}\label{sec:convergence}

In order to carefully analyze the convergence of LiSSA iterations \eqref{lissa_update} we reformulate them as SGD updates. Observe that the result of iHVP applied to a gradient $\gv$, $\uv^{\star} = (H + \lambda)^{-1} \gv$, delivers minimum to the following objective
\begin{equation}\label{quadratic_objective}
    \uv^{\star} = \arg\min_{\uv} L(\uv), \qquad L(\uv) := \frac{1}{2} \uv^{\T} {H} \uv + \frac{\lambda}{2} \| \uv\|^{2} - \uv^{\T} \gv \,.
\end{equation}
With appropriate scaling, the LiSSA updates are equivalent to SGD with step size $\eta$ and the gradient calculated on in-batch loss $\tilde{L}_{t}(\uv)$, where the Gauss-Newton Hessian $H$ is replaced with unbiased estimate $\tilde{H}_t$ calculated over a random batch, turning the updates in \eqref{lissa_update} into
\begin{equation*}
    \uv^{t} = \uv^{t-1} - \eta \left[(\tilde{H}^{t} + \lambda) \uv^{t-1} - \gv \right], \qquad t = 1, \dots, T
\end{equation*}
where the scaling parameter in \eqref{lissa_update} now plays the role of a learning rate.
Convergence of SGD is well studied in the literature, with the recommending step size $\eta$ typically depending on the smoothness of $L(\uv)$, which in our case equals to $\lambda_{\max}(H)$ \cite{bubeck2015convex}. Although in theory, mini-batch SGD is generally considered more efficient than full-batch, in practice mini-batch SGD often performs poorly due to the difference in  smoothness of the population objective $L(\uv)$ and the in-batch objective $\tilde{L}(\uv)$ as pointed out by \cite{tang2020practicality}. Instead, they formulate their bounds in terms of expected smoothness of $\tilde{L}_{t}(\uv)$. We derive the following results that takes into account this difference for the quadratic optimization \eqref{quadratic_objective}.

%In their paper they are focusing on imaging inverse problems, but we can adapt their Theorem 4.3 for our quadratic objective \eqref{quadratic_objective} with slight changes.

\begin{theorem}\label{thm:lissa_convergence}
Suppose, $\eta < 1 / (\lambda_{\max}(H) + \lambda)$. Then, we have convergence in-expectation
\[
    \| \E \uv^{t} - \uv^{\star}\| \leq (1 - \lambda \eta)^{t} \| \uv^{0} - \uv^{\star}\| \,.
\]

Furthermore, assume that $\eta > 0$, $\delta \in (0, 1)$ are such that
\begin{equation}\label{sampling_correctied_condition}
    (1 - \eta (H + \lambda))^{2} + \eta^{2} (\E \tilde{H}_t^{2} - H^{2}) \preceq (1 - \delta) I .
\end{equation}
Then,
\[
    \E \| \uv^{t} - \uv^{\star} \|^2 \leq (1 - \delta)^{t} \left( 2 \| \uv^{0} - \uv^{\star}\|^{2} + \| \uv^{\star}\|^2\right) + \delta^{-1} \eta^{2} \tilde{\Delta},
\]
where we interpret $\tilde{\Delta} = \E \| (H - \tilde{H}_t) \uv^{*}\|^{2}$ as a sampling error.
\end{theorem}

We conclude that although the updates of LiSSA are designed to be unbiased, it is not always guaranteed to converge even if $ \eta < 1 / \lambda_{\max}(H)$. 
In Section~\ref{counter-example}, we show a counter-example where the difference $\E\|\uv^{t} - \uv^{\star}\|^{2}$ is not guaranteed to converge whenever the requirement \eqref{sampling_correctied_condition} does not hold. 
The principal difference comes from the matrix $\E \tilde{H}_t^2 - H^2$, which can be interpreted the sampling gap, i.e. the larger the batch size, the closer the sampled squared Hessian is to the population squared Hessian. In particular, as we increase the batch size $|B|\rightarrow \infty$, sooner or later we expect to have that $\E \tilde{H}_t^2 \approx H^2 $. Thus, the batch size has direct impact on the convergence of LiSSA. Although it is hard to assess the inequality \eqref{sampling_correctied_condition} directly, we find that in many cases we can rely on the following simple condition

% ... (in your document)

\begin{equation}
    \boxed{\E\tilde{H}_t^2 - H^2 \preceq \frac{C}{|B|}\mathrm{Tr}(H)H,}
    \tag{C.1}
    \label{eq:condition1}
\end{equation}

where $ |B| $ is the batch size, which for language models corresponds to the total in-batch \emph{number of tokens}. For simplicity, we assume this number to be the same in each batch $B$. We think of $C $ as a constant moderately larger than $1$, e.g. $C = 2$.

In particular, we find this condition to hold for classification with independent sampling (assuming certain properties of the gradients' distribution). For the case of language modeling, the batch consists of tokens sampled per sequence. We show that the condition holds if the gradients corresponding to different tokens within the same sequence have (mostly) little correlation. This is reasonable to expect due to the large dimension of the parameter, and we also note that a similar assumption appears in \cite{tang2020practicality} in the context of imaging inverse problems.
In addition, we provide a simple empirical test where we compare the traces of LHS and RHS in \eqref{eq:condition1}, and it confirms the inverse scaling with batch size for both classification and language modeling tasks. See Section~\ref{condition_motivation} in the appendix for these derivations and experiments.

% 
%Finally, we conduct two empirical sanity checks: we evaluate the traces of the matrices on both sides of the inequality \eqref{conjecture_matrix_ineq} for different batch sizes, which turn out to be pretty close. \textcolor{red}{We also assess the matrix inequality based on random sketching \citep{needell2022testing} [TBD]}. Although it does not guarantee that the condition holds, it serves as a strong evidence in its favor. We go through each of the above arguments in detail in the Appendix, Section~\ref{} \textcolor{red}{[TBD]}.

Under condition~\ref{eq:condition1}, we can rewrite Theorem~\ref{thm:lissa_convergence}  in a simplified form with an exact requirement for a sufficiently large batch size.

\begin{corollary}\label{corollary_1}
Suppose that~\ref{eq:condition1} holds. Let us choose the hyperparameters
\[
    \eta = 1/ (\lambda_{\max}(H) + \lambda), \qquad |B| \geq C \Tr(H) / \lambda_{\max}(H) 
\]
Then,
\begin{equation}\label{corollary_bound}
    \E \| \uv^{t} - \uv^{*} \|^{2} \lesssim (1 - \lambda \eta)^{2t} (\| \uv^0\|^{2} + \| \uv^{\star}\|^{2})  + \frac{\eta^{2}\Tr(H)}{|B|} \gv^{\T} (H + \lambda)^{-1} \gv
\end{equation}
and the algorithm converges in $T = \Omega(1 / (\eta \lambda))$ steps.
\end{corollary}

In our error bound above, the first term depends on the learning rate and the number of steps, and we can say that it measures how quickly we converged to the solution, therefore we label it \emph{convergence error}. The second term depends directly on the batch size and we label it \emph{sampling error}. It does not depend on the number of steps performed and comes from the difference between the sampled HVP and population HVP. In particular, it is trivial to see that it corresponds to the variance of a single update, $ \E \| \uv^{t} - \E[\uv^{t} | \uv^{t-1}] \|^{2} $, in the limit $\uv^t \rightarrow \uv^{\star}$.  Notice that although the \emph{convergence error} does not depend on the batch size explicitly, we have to satisfy the condition $|B| \geq C \mathrm{Tr}(H)/\lambda_{\max}(H)$ in order to converge. We also note that in terms of batch size requirement our result conforms with \cite{tang2020practicality}. However, they only make a qualitative characterization that mini-batch SGD is applicable in problems where $H$ has fast decaying eigenspectrum.

\begin{remark}
    Notice that Theorem~\ref{thm:lissa_convergence} can hold for arbitrary matrix $H$ and its sampling counterpart, such as the original Hessian $ \E \nabla^{2} \ell(z; \theta) $, and arbitrary unbiased estimates $\tilde{H}_t$, assuming the batches are drawn independently of each other and from the same distribution. However, condition~\eqref{eq:condition1} and, correspondingly, Corollary~\ref{corollary_1} are only expected to hold for the Gauss-Newton Hessian.
\end{remark}

%$\exp(-\eta \lambda T) P = L,  \eta^{2} / |B| Q = L$, $ T|B|^{1/2} = \frac{\sqrt{Q/L}}{\lambda} \log (P / L) $, $ \eta^2 |B| Q = L $ Largest usable correponds to $\eta = 10^{-6}$, so $10^{-6} / |B|^{1/2} = L / P$, $|B| = (L / P) 10^{12}$. Typically $P$ is large large and $L$ is small... So I do not think it contradicts the idea that larger batch sizes is better...

\begin{table}[t]
\centering
\begin{threeparttable}
  \caption{Gauss-Newton-Hessian statistics and recommended hyperparameters. Statistics are calculated on ImageNet (IN) on vision and Open-Web-Text-2 (OWT) on language models.  The arrow $\uparrow$ indicates a lower bound, and  $\downarrow$ indicates that it is an upper bound.}
  \label{hessian-stats-table}
  \begin{tabular}{lrrrrlrr}
    \toprule
    &&&& \multicolumn{1}{c}{$\lambda_{\max}(H)$} & \multicolumn{3}{c}{\makecell{recommended \\ hyperparameters\tnote{*}}}                   \\
    \cmidrule(r){5-5}
    \cmidrule(r){6-8}
    Model     & Size & Data&  $\tfrac{1}{N} \Tr(H)$     & \makecell{from \\ sketching} & \makecell{$\downarrow \eta$} & \makecell{$\uparrow |B|$} & \makecell{$\uparrow  T$} \\
    \midrule
    ResNet-18 & 11M & IN & $(1.32 \pm 0.00) \times 10^{-3}$ & $\approx 270$ & 0.003 & 100 &  150    \\
    ResNet-50 & 25M & IN  & $(8.17 \pm 0.11) \times 10^{-4}$ &  $\approx 470$ & 0.002 & 5 &  200  \\
    OPT &1.3B & OWT    & $(9.28 \pm 0.35) \times 10^{-6}$  & $\approx 780$  & 0.001 & 30 & 500  \\
    Llama-1 & 7B  & OWT   &   $(5.69 \pm 0.67) \times 10^{-6}$     & $\approx 1600$ & 0.0005 & 50 & 1000 \\
    Mistral & 7B & OWT    &   $ (8.18 \pm 0.13) \times 10^{-5}$    & $\approx 5600$ & 0.0002 & 200 & 2000  \\
    \bottomrule
  \end{tabular}
  \begin{tablenotes}
  \item[*]  Assuming $C = 2$ in~\ref{eq:condition1}, and we take $\lambda = 5.0$, $T = 2/(\lambda \eta)$ for sake of demonstration.
  \end{tablenotes}
  \end{threeparttable}
\end{table}

\paragraph{Empirical analysis of eigenvalue statistics.} It is apparent that to choose hyperparameters correctly, we need to evaluate the statistics $\lambda_{\max}(H)$ and $\Tr(H)$. Since there is no way to calculate the Hessian explicitly, we resort to random feature methods that only require evaluation of HVPs.

Evaluating the trace is straightforward. We generate a series of quadratic forms $ (\gv_i^{\T} \tilde{H}_i \gv_{i})_{i = 1}^{N}$, which consists of calculating HVPs and dot products for Gaussian vectors $ \gv \sim \mathcal{N}(0, \tfrac{1}{N} I)$. Their mean estimates the trace \( \E \gv^{\T} \tilde{H} \gv = \E \gv^{\T} H \gv = \frac{1}{N} \Tr(H)\). Due to independence of observations, we can also evaluate standard error of this estimator.
For evaluating the largest eigenvalue we use \emph{random sketching}. That is, we evaluate the matrix $\hat{H} = \Phi H \Phi^{\T}$, with $\Phi \in \R^{d \times N}$ generated in a way such that $ \Phi_{i, j} \sim \mathcal{N}(0, 1/d) $. It is known, such sketches can evaluate the top eigenvalues of the original matrix $ \hat{\lambda}_i(H) = \lambda_i(\hat{H}) - \frac{1}{d} \Tr(\hat{H}) $ \citep{swartworth2023optimal}, with error of estimation negligible for the top eigenvalue. See calculation details in the appendix, Section~\ref{sketching_details}.

We report the results of these evaluations in Table \ref{hessian-stats-table} for 2 ResNets and 3 open-sourced language  models. We also show recommendations for the choice of hyperparameters based on Corollary \ref{corollary_1}. Notice that contrary to the original idea of SGD, in all  cases the recommended batch size is larger than $1$. 
%This observation highlights the gap between typical theoretical assumptions and practical problems.
However, recall that for language modeling, the batch size is the amount of tokens in a batch, and the recommended values are smaller than a typical context length. Thus, the LiSSA can work with just one sequence per batch. We also note that the recommendation is only a lower bound that ensures that LiSSA does not diverge, and increasing the batch size further makes the sampling error smaller (second term in \eqref{corollary_bound}).

%\begin{remark}[On repetition] In order to reduce the variance, previous work suggested averaging the result over repeating the LiSSA for $R > 1$ times. For example, \cite{koh2017understanding} choose $R = 10$ repetitions and \cite{bae2022if} do $R = 5$ repetitions in their experiment. It is not clear however how to choose this yet another hyperparameter. Our result suggests that it is more beneficial to reduce the step size $\eta$ and increase the amount of steps. In fact, we can achieve the same bound with $\sqrt{R}$-times less steps in total compared to repeating, however, we have to sacrifice the ability to do so in parallel. We expand details in the appendix, Section~\ref{sec:repeat}.
%\end{remark}

\section{Empirical validation} % Proximal-Bregman Retraining Function}

\begin{figure}[t]
    \centering
    \includegraphics[width=\textwidth]{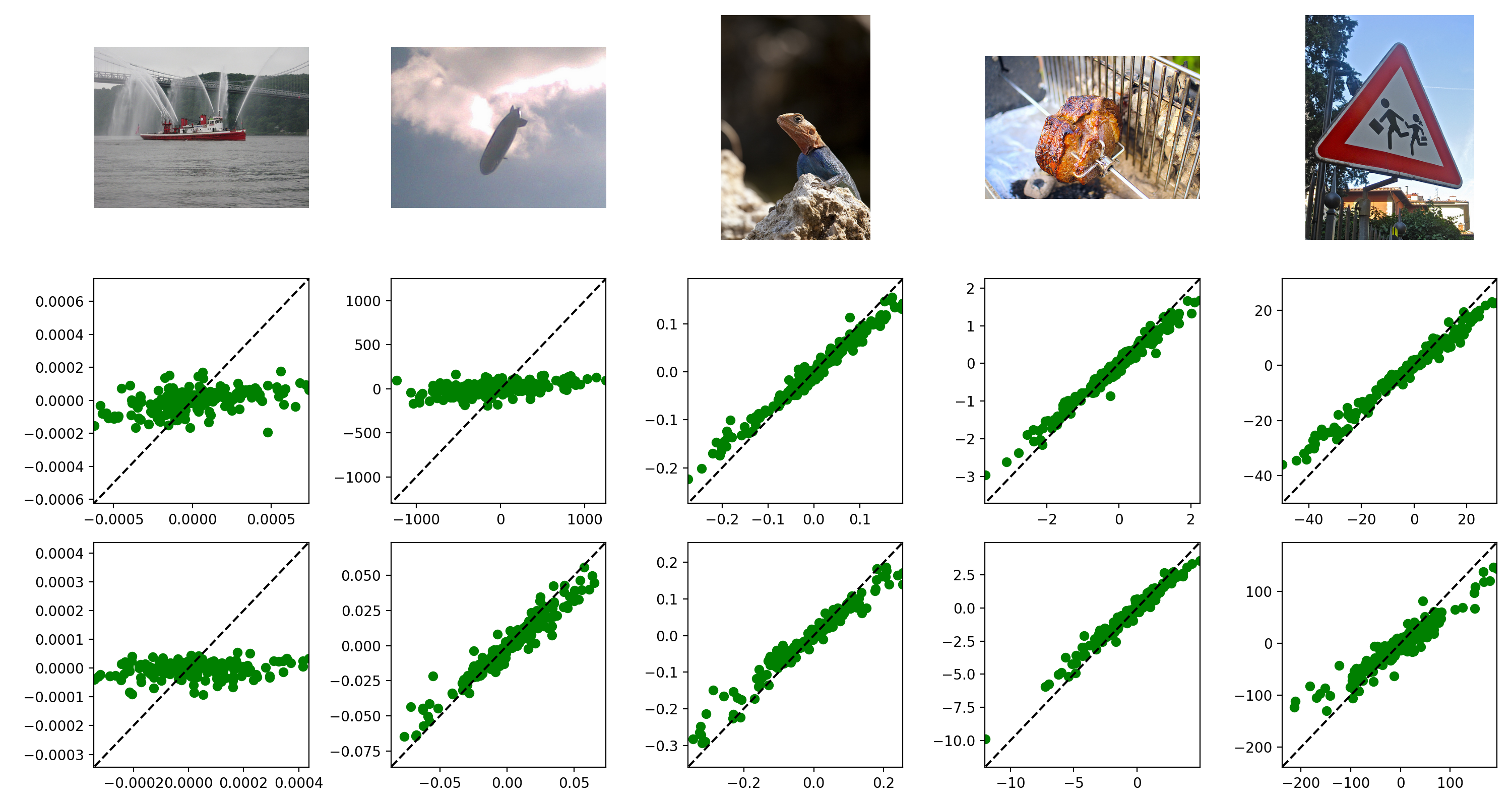}
    \caption{Comparison of PBRF and LiSSA influence. The first row shows examples of training images. Below, the $x$-axis represents LiSSA influences, and the $y$-axis represents the PBRF influences corresponding to each training image and 500 test images. The second row is for ResNet-18, and the third row is for ResNet-50.
    }
    \label{fig:pbrf_vs_lissa}
\end{figure}

\begin{figure}[t]
    \centering
    \includegraphics[width=\textwidth]{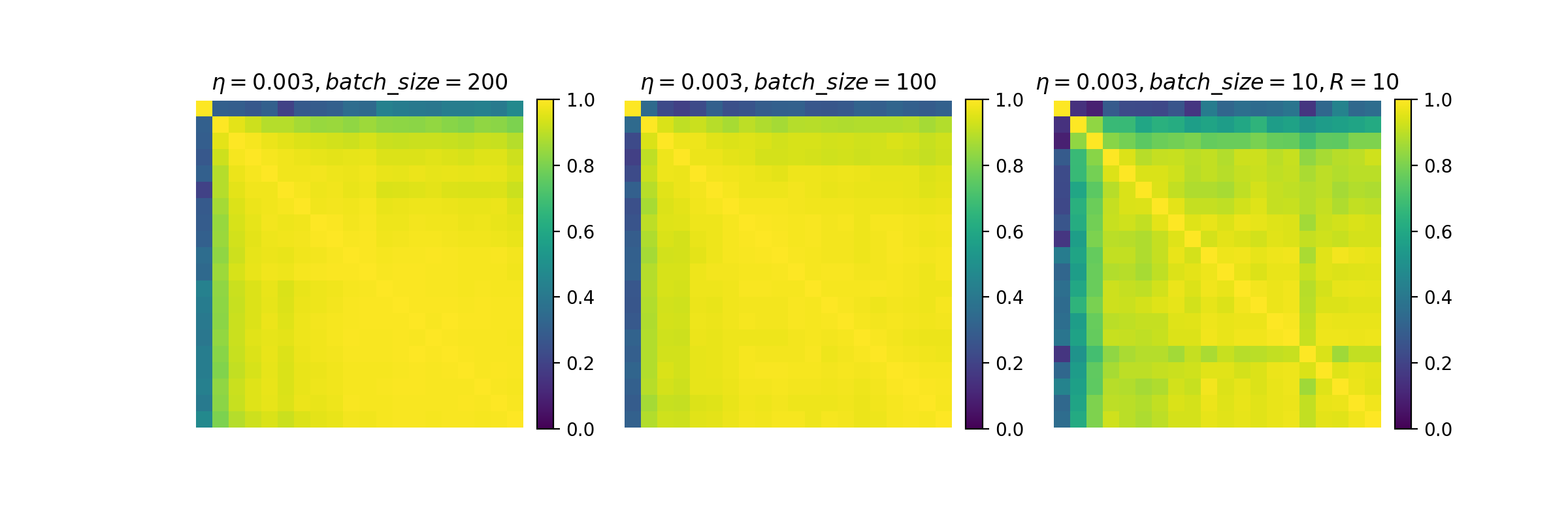}
    \caption{Convergence of LiSSA for ResNet-18 with different batch size configurations. We calculate the correlation between test influences at steps 1..1000 of LiSSA. The result for the small batch size of 10 is averaged over 10 trials, so that the amount of data used in the middle and rightmost figures is the same.}
    \label{fig:convergence}
\end{figure}

We now conduct empirical validation of our theoretical  results. In particular, we want to check two things. Firstly, we want to demonstrate that when the parameters are chosen according to Table~\ref{hessian-stats-table}, the LiSSA converges as expected. For ground truth, we calculate PBRF for selected training examples \cite{bae2022if}. Secondly, we want to empirically confirm that the requirement on sufficiently large batch size is indeed important.

We compare LiSSA and PBRF for the ResNet-18 and ResNet-50 models and randomly selected 25 training and 500 test images from the ImageNet dataset. For each training image we calculate iHVP $ s_{train} = (H + \lambda)^{-1} \nabla \ell(x_{train}, y_{train}) $ using LiSSA with hyperparameters from Table \ref{hessian-stats-table}, and calculate the 500 influences $s_{train}^{\T} \nabla \ell(x_{test}, y_{test})$. We also calculate the PBRF by finetuning the model with SGD on Proximal Bregman objective. For PBRF, we match the batch size, number of steps, and the learning rate to LiSSA. We take $\epsilon =$1e-8 in \eqref{pbrf_equation} and optimize the PBO using double precision to avoid float overflow. We show results for ResNet-18 and ResNet-50 for selected 5 images, and the full list is shown in the appendix, Section \ref{sec:appendix_pbrf}. We observe three cases: 1) LiSSA approximates PBRF, i.e. scatter plot concentrates along the dashed line $x = y$; 2) both LiSSA and PBRF have very low values poorly distinguishable from zero; 3) both LiSSA and PBRF have high value and do not approximate each other. In the latter case, we can argue that the PBO finetunning stirs away the model too far for the quadratic approximation to hold.

Furthermore, we confirm that the batch size matters not only for the sampling error in \eqref{corollary_bound}, but also for the speed of convergence. Let us take ResNet-18 with damping parameter and run the LiSSA algorithm for 1000 steps. According to Table \ref{hessian-stats-table}, the recommended batch size is equal to 100. We suggest to consider three set-ups: batch size 2x larger than recommended, batch size is equal to 100, and batch size 10x smaller than recommended, with the result averaged over 10 independent runs. In the latter and the former cases, the result is obtained through iterating in the same amount of data, which allows to equalize the sampling error of a single update. Figure~\ref{fig:convergence} reports the correlation between influences calculated for 500 test images at different steps. As we can see the correlation converges to 1 faster for the two cases where the batch size is greater or equal to recommended 100, despite averaging over 10 trials for the small batch.

\section{What is the role of inverted Hessian?}\label{sec:hess_role}

In the context of language models, the focus in the current literature is mostly on gradient-based influence functions \citep{xia2024less, he2024s, chhabra2024outlier}. Often this choice is motivated by simpler and faster implementation. Due to the high cost of Hessian-based influence calculation, it is natural to ask what are the benefits compared to the gradient-based influence. We conduct a simple experiment in an attempt to understand what is left out of the consideration when relying only on gradient dot products.

Consider the eigenvalue decomposition of the Gauss-Newton Hessian $H$,
\[
    H = \sum_{j = 1}^{N} \lambda_j v_j v_j^{\T},
\]
where $v_j$ are orthogonal and normalized and $H v_j = \lambda_j v_j$. If we represent a gradient $g = \nabla \ell(z_{test})$ in this eigenbasis, the iHVP simply reweights the coefficient according to how large the eigenvalue is,
\begin{equation}\label{inverse_hessian_eigendecomposition}
    g = \sum_{j} \langle g, v_j \rangle v_j,
    \qquad
    \lambda (H + \lambda)^{-1} g = \sum_{j} \frac{\lambda }{\lambda_j + \lambda} \langle g, v_j \rangle v_j \,.
\end{equation}
It is known that for classification tasks (what language models do per token), the Gauss-Newton Hessian is equivalent to a form of variance of the generated gradients $\nabla \ell(\hat{y} | x)$, where $\hat{y} \sim p(y|x)$, which is referred to as Fisher Information Matrix (FIM). Generally speaking this is different from the empirical FIM $\E_{z_{train} \sim \mathcal{D}_{train}} \nabla \ell(z_{train}) \nabla \ell(z_{train})^{\T} $, however, for low noise distributions the two might be used interchangeably \citep{martens2020new}. Such interpretation can help us to speculate, that the directions $v_j$ corresponding to higher eigenvalues $\lambda_j$ are more likely to observe in the training gradients, in the sense that $\E \langle g, v_j \rangle^{2}$ is higher. We also notice that $ \frac{\lambda}{\lambda + \lambda_j} \approx 0 $ for the top eigenvalues $\lambda_j$ that are much larger than the damping parameter $\lambda$. On the contrary, the lower eigenvalues receive higher weight, that is $ \frac{\lambda}{\lambda + \lambda_j} \rightarrow 1 $ as $\lambda_j \rightarrow 0$. In this sense, the iHVP works contrary to the traditional Principal Component Analysis, where the idea is to project the vector onto the top eigenvectors of the covariance. Instead, applying inverse Hessian removes the top directions corresponding to $\lambda_j \gg \lambda$ and retains the directions corresponding to $\lambda_j \ll \lambda$.

%\begin{remark}
%    On TF-IDF
%\end{remark}

\begin{table}[t]
\caption{Examples of original and paraphrased sentences.}
\label{tab:sentence_examples}
\centering
\begin{tabularx}{0.9\textwidth}{>{\hsize=0.3\hsize}X|>{\hsize=1.7\hsize}X}
\toprule
{original} & {\it ‘‘The Mona Lisa, painted by Leonardo Da Vinci in the early 16th century, is one of the most famous paintings in the world.''} \\
\hline
{paraphrased} & {\it ‘‘Leonardo Da Vinci's early 16th-century painting, the Mona Lisa, is widely regarded as one of the most renowned artworks globally.''} \\
%\hline
%{fake} & {\it ‘‘The Mona Lisa, painted by Leonardo DiCaprio in the early 16th century, is actually a self-portrait of the actor wearing a wig and a dress.''}
%\\
\bottomrule
\end{tabularx}
\end{table}

For example, a plausible interpretation of the directions $v_j$ would be that the top directions correspond to general language coherence, sentence structure, and keywords, while the directions with small eigenvalues could correspond to more specific, informative content.
We propose the following experiment to encourage such point of view. We consider ten pairs of sentences, one related to some historical or scientific fact, we refer to as \emph{original}, the other is a paraphrased version of the same fact,
referred to as  \emph{paraphrased}.
%The third sentence changes the original to make it a completely made up fact, but keeps the structure of the sentence, referred to as \emph{fake}.
We show one such pair in Table~\ref{tab:sentence_examples} and in Appendix~\ref{similarity_prompts} we give all 10 pairs\footnote{To avoid cherry-picking, all 20 sentences were generated with Claude 3 Opus with a few prompts.}.
For each of the sentences, we calculate the gradient of the next word prediction loss $\nabla \ell(z)$ and calculate pairwise dot-influences $ \nabla \ell(z)^{\T} \nabla \ell(z') $ and Hessian-based influences $ \nabla \ell(z)^{\T} (H + \lambda)^{-1} \nabla \ell(z') $.
Our goal is to measure the similarities between original sentences, their rewritings, and their made-up derivatives. For this, we propose to measure the similarity by correspondingly normalizing with norms of gradients and self-influence:
\begin{align*}
    \text{\rm Gradient-similarity}(z, z') &= \frac{\nabla \ell(z)^{\T} \nabla \ell(z')}{\|\nabla \ell(z)\| \| \nabla \ell(z')\| }, \\
    \text{\rm Influence-similarity}(z, z') &= \frac{\nabla \ell(z)^{\T} (H + \lambda)^{-1} \nabla \ell(z')}{\sqrt{\nabla \ell(z)^{\T} (H + \lambda)^{-1} \nabla \ell(z)}\sqrt{\nabla \ell(z')^{\T} (H + \lambda)^{-1} \nabla \ell(z')}} .
\end{align*}

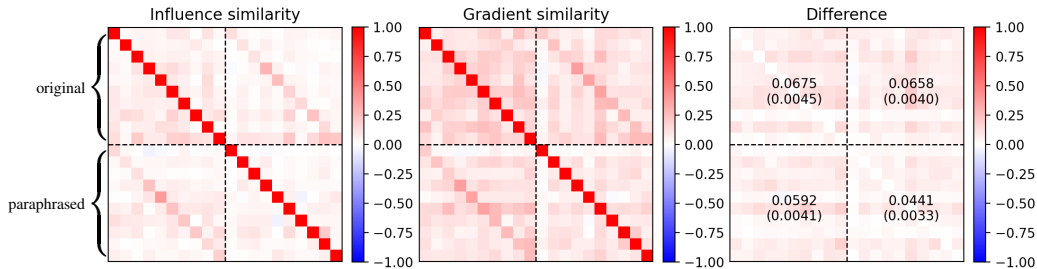
\begin{figure}[t]
    \centering
    %\phantom{AA}
    %\begin{subfigure}{0.9\textwidth}
    %    \includegraphics[width=\textwidth]{figures/simfig_1.png}
    %    \begin{picture}(0,0)
    %        \put(-20,110){\rotatebox{0}{\text{\tiny original}}}
    %        \put(-30,81){\rotatebox{0}{\text{\tiny paraphrased}}}
    %        \put(-11.5,52){\rotatebox{0}{\text{\tiny fake}}}
    %        \put(-8,67){\rotatebox{270}{\text{$\underbrace{\hspace{.95cm}}_{}$}}}
    %        \put(-8,96){\rotatebox{270}{\text{$\underbrace{\hspace{.95cm}}_{}$}}}
    %        \put(-8,125){\rotatebox{270}{\text{$\underbrace{\hspace{.95cm}}_{}$}}}
    %    \end{picture}
    %\end{subfigure}
    \phantom{AA}
    \begin{subfigure}{0.9\textwidth}
        \includegraphics[width=\textwidth]{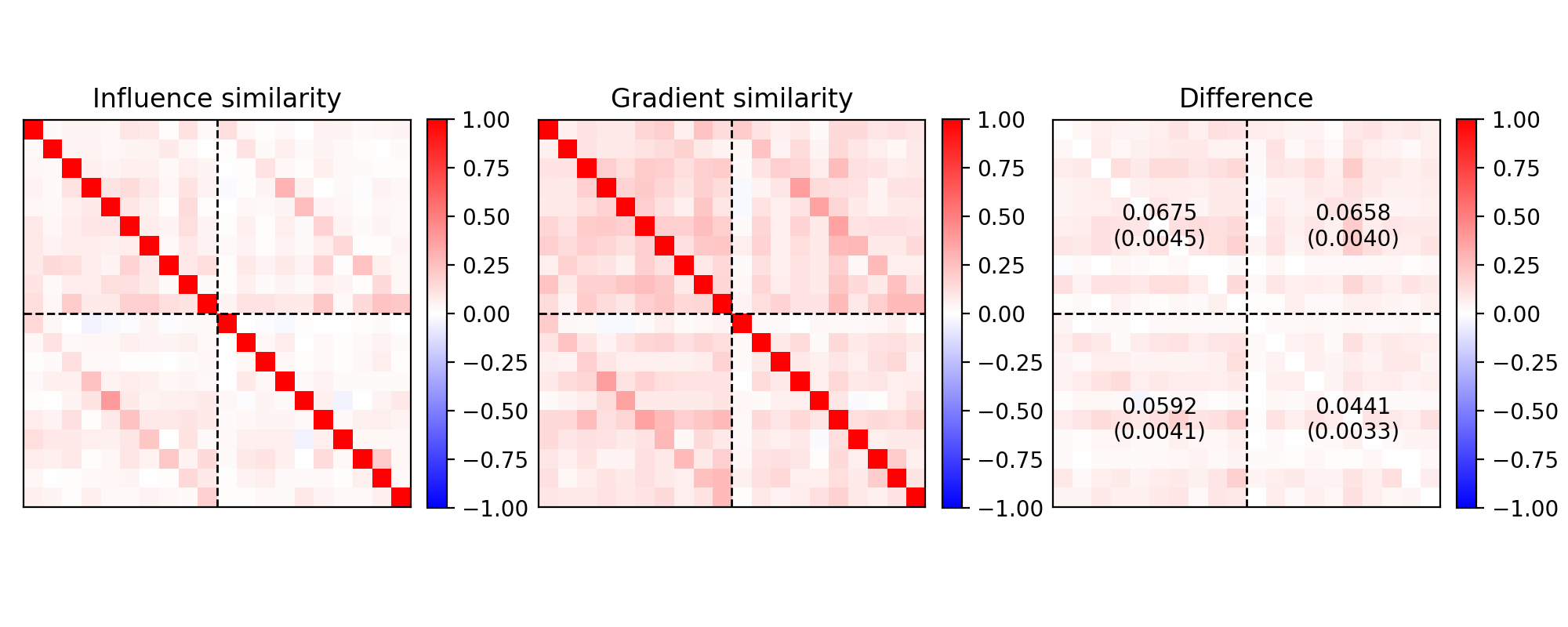}
        \begin{picture}(0,0)
            \put(-22,102){\rotatebox{0}{\text{\tiny original}}}
            \put(-32.5,58){\rotatebox{0}{\text{\tiny paraphrased}}}
            \put(-8,80){\rotatebox{270}{\text{$\underbrace{\hspace{1.4cm}}_{}$}}}
            \put(-8,124){\rotatebox{270}{\text{$\underbrace{\hspace{1.4cm}}_{}$}}}  
        \end{picture}
    \end{subfigure}
    \caption{Similarity between 20 sentences, see complete list in in Appendix, Section~\ref{similarity_prompts}. Left figure shows influence similarity calculated with LiSSA, middle  --- gradient similarity, right --- the difference between the former and the latter. In the rightmost figure the numbers show the mean over each 10x10 square, with standard error in the brackets. We use OPT 1.3B model, with $\lambda = 5.0$, $T = 1000$ and $\eta = 0.003$. We also use batch size of $4$ sequences, each consisting of $512$ tokens.}
    \label{fig:sim}
\end{figure}

We show the pairwise similarities between all 20 sentences in Figure~\ref{fig:sim}. In the rightmost graph we also show the difference between gradient similarity and influence similarity. We observe that unrelated sentences generally have higher gradient similarity than influence similarity since the values in the rightmost graph are mostly positive.
As a result, the influence similarity between an original sentence and a rewritten one appears to be consistently higher than between unrelated sentences. %On the other hand, both influence and gradient similarities capture the lack of connection between sentences from different modalities --- correct and fake ones. We also observe a mildly lower influence similarity between unrelated fake sentences. These observations confirm the idea described above that Hessian inverse downweights the information that can be observed in any sentence that e.g. contains factual information.

Downweighting directions that are more likely to observe in \eqref{inverse_hessian_eigendecomposition} can also be compared to the idea of the TF-IDF index, where the terms are reweighted according to their inverse frequency \citep{salton1983introduction}. Incidentally, we show that for a bag-of-words model (which although trivial, is also a language model), the influence functions correspond to a particular form of the TF-IDF index, see Section~\ref{section-tf-idf} in the Appendix. %We also note that for larger $\lambda$, which downweights less eigenvectors in \eqref{inverse_hessian_eigendecomposition}, we require smaller number of steps according to Corollary~\ref{corollary_1}, therefore we need to ``see'' less information.

%\section{Related work}

\section{{Conclusion}}

We have shown that the hyperparameters of the classical LiSSA algorithm  can be chosen based on two spectral statistics of the Gauss-Newton Hessian: its trace and largest eigenvalue. In particular, we show the batch size used for sampling  Hessian-vector products per update has to be sufficiently large. Otherwise, LiSSA might not converge which we demonstrate empirically and theoretically. This particular aspect of hyperparameter choice for LiSSA algorithm has not been previously addressed in the literature. Furthermore, we empirically demonstrate that applying to large models in its original form can still be feasible if we choose a sufficiently large damping parameter. 

We do not necessarily advocate for using LiSSA as the algorithm of choice, rather it can be used as a baseline for validating other more lightweight algorithms, in which case it is important to make sure that the hyperparameters of LiSSA are chosen correctly. We hope that our result and the implementation can further facilitate research in influence functions, as well as in other topics where inverse Hessian-vector products naturally appear \citep{guo2019certified, Schulman2015TrustRP, martens2020new, kirkpatrick2017overcoming}.

\bibliography{main}

\appendix

\section{List of prompts for the experiment in Section \ref{sec:hess_role}}\label{similarity_prompts}

\begin{enumerate}
\item \phantom{a}
\begin{center}
\begin{tabular}{p{1.2cm}|p{10cm}}
{original} & {\it ``The Great Wall of China is the longest wall in the world, stretching over 21,000 kilometers.''}\\
{rewrite} & {\it ``Spanning over 21,000 kilometers, the Great Wall of China holds the record for being the longest wall worldwide.''}\\
%{fake} & {\it ``The Great Wall of China is a tiny fence in Mongolia, stretching over 21 centimeters.''}\\
\end{tabular}
\end{center}
\item \phantom{a}
\begin{center}
\begin{tabular}{p{1.2cm}|p{10cm}}
{original} & {\it ``In 1969, Neil Armstrong became the first human to set foot on the Moon during the Apollo 11 mission.''}\\
{rewrite} & {\it ``During the Apollo 11 mission in 1969, Neil Armstrong made history by becoming the first person to walk on the lunar surface.''}\\
%{fake} & {\it ``In 1969, Darth Vader became the first Sith Lord to set foot on the Moon during the Apollo 11 mission, which was actually a Galactic Empire propaganda stunt.''}\\
\end{tabular}
\end{center}
\item \phantom{a}
\begin{center}
\begin{tabular}{p{1.2cm}|p{10cm}}
{original} & {\it ``The theory of evolution by natural selection was first proposed by Charles Darwin in his book "On the Origin of Species" in 1859.''}\\
{rewrite} & {\it ``Charles Darwin introduced the concept of evolution by natural selection in his 1859 publication titled "On the Origin of Species".''}\\
%{fake} & {\it ``The theory of evolution by natural selection was first proposed by Captain Jack Sparrow in his book "On the Origin of Rum" in 1859, after a wild night in Tortuga.''}\\
\end{tabular}
\end{center}
\item \phantom{a}
\begin{center}
\begin{tabular}{p{1.2cm}|p{10cm}}
{original} & {\it ``The United Nations was founded in 1945 after World War II to maintain international peace and security.''}\\
{rewrite} & {\it ``Following the conclusion of World War II, the United Nations was established in 1945 to foster international peace and security.''}\\
%{fake} & {\it ``The United Nations was founded in 1945 after World War II to organize the best global pizza party every year.''}\\
\end{tabular}
\end{center}
\item \phantom{a}
\begin{center}
\begin{tabular}{p{1.2cm}|p{10cm}}
{original} & {\it ``The Eiffel Tower, constructed in 1889 for the World's Fair, is one of the most iconic landmarks in Paris, France.''}\\
{rewrite} & {\it ``One of the most recognizable structures in Paris, France, the Eiffel Tower was built in 1889 for the World's Fair.''}\\
%{fake} & {\it ``The Eiffel Tower, also known as the Giant Jenga Tower, was accidentally constructed in 1889 when Gustave Eiffel sneezed during a game of Jenga in Paris.''}\\
\end{tabular}
\end{center}
\item \phantom{a}
\begin{center}
\begin{tabular}{p{1.2cm}|p{10cm}}
{original} & {\it ``The French Revolution, which began in 1789, marked the end of the monarchy and the establishment of a republic in France.''}\\
{rewrite} & {\it ``The monarchy in France was abolished, and a republic was established as a result of the French Revolution, which commenced in 1789.''}\\
%{fake} & {\it ``The French Revolution, which began in 1789, was actually a misunderstanding over a croissant recipe that escalated quickly.''}\\
\end{tabular}
\end{center}
\item \phantom{a}
\begin{center}
\begin{tabular}{p{1.2cm}|p{10cm}}
{original} & {\it ``The human brain contains approximately 86 billion neurons, making it the most complex organ in the human body.''}\\
{rewrite} & {\it ``The most intricate organ in the human body, the brain, is composed of roughly 86 billion neurons.''}\\
%{fake} & {\it ``The human brain contains approximately 86 billion tiny aliens, making it the most complex organ in the universe, according to conspiracy theorists.''}\\
\end{tabular}
\end{center}
\item \phantom{a}
\begin{center}
\begin{tabular}{p{1.2cm}|p{10cm}}
{original} & {\it ``In the 2020 United States presidential election, Joe Biden defeated incumbent Donald Trump to become the 46th president.''}\\
{rewrite} & {\it ``Joe Biden secured victory over the sitting president, Donald Trump, in the 2020 United States presidential election, becoming the 46th president.''}\\
%%{fake} & {\it ``In the 2020 United States presidential election, Kanye West defeated both Joe Biden and Donald Trump by promising every American a free pair of Yeezy shoes.''}\\
\end{tabular}
\end{center}
\item \phantom{a}
\begin{center}
\begin{tabular}{p{1.2cm}|p{10cm}}
{original} & {\it ``The Mona Lisa, painted by Leonardo da Vinci in the early 16th century, is one of the most famous paintings in the world.''}\\
{rewrite} & {\it ``Leonardo da Vinci's early 16th-century painting, the Mona Lisa, is widely regarded as one of the most renowned artworks globally.''}\\
%%{fake} & {\it ``The Mona Lisa, painted by Leonardo DiCaprio in the early 16th century, is actually a self-portrait of the actor wearing a wig and a dress.''}\\
\end{tabular}
\end{center}
\item \phantom{a}
\begin{center}
\begin{tabular}{p{1.2cm}|p{10cm}}
{original} & {\it ``Climate change is a global issue caused by the increase of greenhouse gases in the atmosphere, primarily due to human activities.''}\\
{rewrite} & {\it ``The primary cause of climate change, a worldwide problem, is the accumulation of greenhouse gases in the atmosphere, largely attributed to human activities.''}\\
%%{fake} & {\it ``Climate change is a hoax created by penguins who want to take over the world by melting the ice caps and flooding human cities.''}\\
\end{tabular}
\end{center}
\end{enumerate}

\section{Postponed proofs from Section~\ref{sec:convergence}}

\subsection{Proof of Theorem~\ref{thm:lissa_convergence}}

We write,
\begin{align*}
    \uv^{t} - \uv^{\star} &= \left( (1 - \lambda \eta) I + \eta \tilde{H}_t\right) \uv^{t-1} + \eta \gv \\
    &= \left( I - \eta (\tilde{H}_t + \lambda)\right) (\uv^{t} - (H + \lambda)^{-1} \gv) + \eta (H - \tilde{H}_t) (H + \lambda)^{-1} \gv \\
    &= \left( I - \eta (\tilde{H}_t + \lambda)\right) (\uv^{t-1} - \uv^{\star}) + \eta (H - \tilde{H}_t) \uv^{\star}\,.
\end{align*}
First, taking the expectation and using the fact that $\tilde{H}_t$ and $\uv^t$ are independent,
\begin{align*}
    \E \uv^{t} - \uv^{\star} &= (I - \eta (H + \lambda)) (\E \uv^{t-1} - \uv^{\star}) \\
    &= (I - \eta (H + \lambda))^{t} (\uv^{0} - \uv^{\star}),
\end{align*}
and under $ \eta < 1 / \lambda_{\max}(H) $ the matrix $I - \eta (H + \lambda) \preceq (1 - \lambda \eta) I$ is a contraction, thus the first bound follows.

For the second part, denote $R = \E(I - \eta(\tilde{H}_t + \lambda))^2$. Let us take the conditional expectation of the square norm, conditional on all the sampling before step $t$. Setting  $\mathcal{F}_{t-1} = \sigma(\tilde{H}_1, \dots, \tilde{H}_{t-1})$, we have that 
\begin{align*}
    \E \left[ \| \uv^{t} - \uv^{\star}\|^{2} \vert \mathcal{F}_{t-1} \right] &= (\uv^{t-1} - \uv^{\star})^{\T} \left[\E (I - \eta(\tilde{H}_t + \lambda))^{2} \right](\uv^{t-1} - \uv^{\star}) \\
    &\phantom{=|} + 2 \eta (\uv^{t-1} - \uv^{\star})^{\T} \E (I - \eta (\tilde{H}_t + \lambda)) (H - \tilde{H}^{t}) \uv^{\star} \\
    &\phantom{=|} + \eta^{2} \E \| (H - \tilde{H}_t) \uv^{\star}\|^{2} \\
    & = \| R^{1/2} (\uv^{t-1} - \uv^{\star})\|^{2} + 2 \eta^{2} (\uv^{t-1} - \uv^{*})^{\T} \{ \E \tilde{H}_t^2 - H^{2}\} \uv^{\star} \\
    &\phantom{=|} + \eta^{2} [\uv^{\star}]^{\T} \{ \E \tilde{H}_t^2 - H^{2}\} \uv^{\star} \\
    &\leq (1 - \delta) \| \uv^{t-1} - \uv^{\star}\|^{2} + 2 \eta (\uv^{t-1} - \uv^{\star})^{\T} \{ \E \tilde{H}_t^2 - H^{2}\} \uv^{\star} + \eta^{2} \tilde{\Delta} \, .
\end{align*}
Here, we have used the fact that by our assumption $\E (I - \eta (\tilde{H_t} + \lambda))^{2} \preceq (1 - \delta) I$ is a contraction, since we have that
\begin{align*}
    \E (I - \eta (\tilde{H_t} + \lambda))^{2} &= 
    (1 - \lambda \eta)^{2} I - 2 \eta (1 - \lambda \eta) 
    \E \tilde{H}_t + \eta^{2} \E \tilde{H}_t^{2} \\
    &= (1 - \lambda \eta)^{2} I - 2 \eta (1 - \lambda \eta) 
    {H} + \eta^{2} {H}^{2} \\
    &\phantom{=|} + \eta^{2} (\E \tilde{H}_t^{2} -H^{2}) \\
    &= (1 - \eta (H + \lambda))^{2} + \eta^{2} (\E \tilde{H}_t^{2} - H^{2})\,.
\end{align*}
Taking the unconditional expectation, we obtain that
\begin{align*}
    \E \| \uv^{t} - \uv^{\star}\|^{2} & \leq (1 - \delta) \E \| \uv^{t-1} - \uv^{\star}\|^{2} + 2 \eta^{2} (\uv^{0} - \uv^{*})^{\T} (I - \eta(H + \lambda))^{t-1} \{ \E \tilde{H}_t^2 - H\} \uv^{\star} + \eta^{2} \tilde{\Delta} \\
    &\leq \dots \\
    & \leq (1 - \delta)^{t} \| \uv^{0} - \uv^{\star}\|^{2} + \eta^{2}\tilde{\Delta} \left(1 + (1 - \delta) + \dots + (1 - \delta)^{t - 1}\right) \\
    & \phantom{=} + 2\eta^{2}(\uv^{0} - \uv^{\star})^{\T} \left\{ \sum_{k=0}^{t-1} (1 - \delta)^k (I - \eta(H + \lambda))^{t-k} \right\} \{ \E \tilde{H}_{t}^2 - H^2\} \uv^{\star} \\
    & \leq (1 - \delta)^{t} \| \uv^{0} - \uv^{\star}\|^{2} +  \frac{\eta^{2}\tilde{\Delta}}{1 - (1 - \delta)} \\
    & \phantom{=} + 2\eta^{2}(\uv^{0} - \uv^{\star})^{\T} \left\{ (1 - \delta)^{t} I - (I - \eta(H + \lambda))^{t} \right\} (\eta (H + \lambda) - \delta I)^{-1} \{ \E \tilde{H}_{t}^2 - H^2\} \uv^{\star}\,.
\end{align*}
We apply the Cauchy-Schwartz inequality to the last term. By \eqref{sampling_correctied_condition}, we have that $ \eta^{2} \{ \E \tilde{H}_t^{2} - H^{2} \} \preceq \eta(H + \lambda) - \delta I$.
Thus,
\begin{multline*}
    \left\| \left\{ (1 - \delta)^{t} I - (I - \eta(H + \lambda))^{t} \right\}^{1/2} (\eta (H + \lambda) - \delta I)^{-1} \eta^{2} \{ \E \tilde{H}_{t}^2 - H^2\} \uv^{\star} \right\|^{2}  \\
    \leq (1-\delta)^{t} \left\|  (\eta (H + \lambda) - \delta I)^{-1} \eta^{2} \{ \E \tilde{H}_{t}^2 - H^2\} \uv^{\star} \right\|^{2}
    \leq (1 - \delta)^{t} \| \uv^{\star}\|^{2}.
\end{multline*}
We also have,
\[
    \left\| \left\{ (1 - \delta)^{t} I - (I - \eta(H + \lambda))^{t} \right\}^{1/2} (\uv^{0} - \uv^{\star})\right\|^{2} \leq (1 - \delta)^{t} \| \uv^{0} - \uv^{\star}\|^{2}.
\]
Collecting everything together,
\[
    \E \| \uv^{t} - \uv^{\star}\|^{2} \leq (1 - \delta)^{2} ( \| \uv - \uv^{\star}\|^{2} + \| \uv - \uv^{\star}\| \| \uv^{\star}\|) + \delta^{-1}\eta^{2} \tilde{\Delta}\,.
\]
%For the case where $\lambda_1 > 1$, we use that
%\[
%    \E \| \uv^{t} - \uv^{\star} \|^{2} \geq \left\| \left[\E (I - \eta(\tilde{H}_t + \lambda))^{2} \right]^{1/2} (\uv^{t-1} - \uv^{\star})\right\|^{2} 
%\]

\subsection{Proof of Corollary~\ref{corollary_1}}

We want to show that with $ \frac{C\Tr(H)}{|B|} \geq \eta^{-1} - \lambda $, equation \eqref{sampling_correctied_condition} takes place with $ \delta = 2 \eta \lambda - (\eta \lambda)^{2} $. Denote $K =\frac{C\Tr(H)}{|B|}$. Since Condition~\ref{eq:condition1} holds, we need to show
\[
    (I - \lambda(\eta + H))^{2} + K \eta^{2} H \preceq (1 - \delta) I
\]
The LHS of the above display has eigenvalues $ (1 - \lambda(\eta + \lambda_j))^{2} + K \eta^{2} \lambda_j $, where $\lambda_j$ are the eigenvalues of $H$. It is therefore sufficient to show that
\[
    \max_{a \in[0, \lambda_{\max}]} (1 - \eta(\lambda + a))^{2} + K \eta^{2} a \leq 1 - \delta
\]
We rewrite this condition as
\[
    \max_{a \in [\lambda, \lambda + \lambda_{\max}]} -2 \eta a + \eta^2 a^2 + K \eta^{2} (a - \lambda) \leq - \delta
\]
The minimum of the quadratic function is attained at $ \bar{a} = \eta^{-1} - K / (2\eta) $. In the case where $\bar{a}$ is in the right half of the interval $[\lambda, \lambda + \lambda_{\max}]$, the maximum of the quadratic function is attained at the point $a = \lambda$. This condition rewrites as $ \eta^{-1} - K / (2\eta) \geq \lambda + \lambda_{\max} / 2 $, which using $\eta(\lambda + \lambda_{\max}) = 1$ translates into $ K \geq \eta^{-1} - \lambda $. Thus, under the assumption that the batch size is at least $ |B| \geq \frac{C\Tr(H)}{\eta^{-1} - \lambda} $, we have that
\[
    \max_{a \in[0, \lambda_{\max}]} (1 - \eta(\lambda + a))^{2} + K \eta^{2} a = (1 - \eta \lambda)^{2} \leq 1 - \delta,
    \qquad
    \delta = 2\eta\lambda - (\eta \lambda)^{2}.
\]
With such $\lambda$ it holds $ (1 - \delta)^{t} = (1 - \eta\lambda)^{2t}$. It is left to notice that
\begin{align*}
    \eta^{2} \| \E (\tilde{H}_t - H) \uv^{\star} \|^{2} &= \eta^{2} [\uv^{\star}]^{\T} \{ \E \tilde{H}_t^{2} - H^{2} \} \uv^{\star} \\
    &\leq \frac{(1 + c)\Tr(H)}{|B|} \gv^{\T}(H + \lambda)^{-1} H (H + \lambda)^{-1} \gv \\
    &\leq  \frac{(1 + c)\Tr(H)}{|B|} \gv^{\T}(H + \lambda)^{-1} \gv \,.
\end{align*}

\subsection{Motivation for Condition \ref{eq:condition1}}\label{condition_motivation}

\paragraph{Case of classification.} We first consider the case where the observations in each batch are drawn independently, which could either be the case of classification problem (e.g. on images) or the case where we sample the tokens independently from the set of all training tokens (of course, this would be inefficient and we only consider this case for sake of demonstration).

\def\gv{\mathbf{g}}
\def\xv{\mathbf{x}}
\def\sv{\mathbf{s}}
Consider the gradients
\[
    \gv = \nabla \ell(\hat{y}|\xv),
    \qquad
    \xv \sim \mathcal{D}_{tr},\;\;
    \hat{y} | \xv \sim p(\xv).
\]
Here, $\xv \sim \mathcal{D}_{tr}$ either means that $\xv$ is drawn uniformly from a finite training dataset, or from some population distribution. The expression $\hat{y} | \xv \sim p(\xv)$ means that once we have drawn $\xv$, we then generate a random label $\hat{y}$ according to the model's output distribution $p(\xv)$. For language models, $\xv = (s_1, \dots, s_{t-1})$ is the context and $ \hat{y} $ predicts $s_{t}$. For simplicity, we assume that $\gv$ are spherically distributed,  meaning that there are independent $ r $, $\sv$ such that $ \E r^2 = 1 $ and $\sv$ is distributed uniformly over a unit sphere in $\R^N$, and $  \gv \overset{d}= \sqrt{N} r H^{1/2} \sv$. This helps to account for heavy tails of the gradients that are often observed in practice, which would not be captured, for example, by sub-Gaussian distributions.

The in-batch Gauss-Newton Hessian then reads as
\[
    \tilde{H}_t = \frac{1}{|B|} \sum_{\xv \in B} \E_{\hat{y} \sim p(\xv)} \nabla \ell(\hat{y}|\xv) \nabla \ell(\hat{y}|\xv)^{\T}
\]
Thanks to the fact that the elements in $B$ are i.i.d. we have that
\begin{align*}
    \E \tilde{H}_t^2 =\, &\frac{1}{|B|} \E \sum_{\xv \neq \xv'} \E_{\hat{y} \sim p(\xv)} \nabla \ell(\hat{y}|\xv) \nabla \ell(\hat{y}|\xv)^{\T} \E_{\hat{y} \sim p(\xv')} \nabla \ell(\hat{y}|\xv') \nabla \ell(\hat{y}|\xv')^{\T} \\
    &+ \frac{1}{|B|^2} \sum_{\xv \in B} \left[\E_{\hat{y} \sim p(\xv)} \nabla \ell(\hat{y}|\xv) \nabla \ell(\hat{y}|\xv)^{\T}\right]^2 \\
    =\, & (1 - 1/|B|) H^2 + \frac{1}{|B|^2} \sum_{\xv \in B} \left[\E_{\hat{y} \sim p(\xv)} \nabla \ell(\hat{y}|\xv) \nabla \ell(\hat{y}|\xv)^{\T}\right]^2 \\
    \preceq\, &
    (1 - 1/|B|) H^2 + \frac{1}{|B|} \E \| \nabla \ell(\hat{y}|\xv)\|^{2} \nabla \ell(\hat{y}|\xv) \nabla \ell(\hat{y}|\xv)^{\T}
\end{align*}
Now, writing $\gv $ instead of $\nabla \ell(\hat{y}|\xv)$ and using the spherical property $\gv = \sqrt{N} r H^{1/2}\sv$, we have that
\[
    v^{\T} [\E \|  \gv\|^{2} \gv\gv^{\T}]v = N^{2} \E r^{4} \E \sv^{\T} H \sv (\sv^{\T} H^{1/2} vv^{\T} H^{1/2} \sv)
\]
For $\sv$ uniformly distributed over a unit sphere, we can use the identity $ \E (\sv^{\T} A \sv)(\sv^\T B \sv) = \frac{1}{N(N+2)} (2 \Tr(AB) + \Tr(A)\Tr(B)) $ \citep{wiens1992moments}. Then, we have that
\[
    v^{\T} [\E\|\gv\|^{2}\gv\gv^{\T}]v = (\E r^4) \frac{N}{N+2} (2 v^{\T}H^2v + \Tr(H) v^{\T} H v),
\]
therefore,
\[
    \E \tilde{H}_t^2 - H^2 \preceq \frac{1}{|B|}  \left\{ (\E r^{4}) \Tr(H) H  + (\E r^{4} - 1) H^2 \right\}.
\]

\paragraph{Case of incoherent tokens.}
Let us now consider the case where we sample the tokens sequence-wise. For simplicity, we assume that each sequence has the same size, so that each token in the dataset has equal probability to be drawn (when we say token to be drawn, we mean that we consider the prediction of token $s_t$ with context $(s_1, \dots, s_{t-1})$. This means that when $s \in B$, and $|s| = L$, our batch contains all contexts $ (s_1, \dots, s_{t-1}) $ for $t = 1, \dots, L$. Here, we assume that gradients corresponding to tokens in the same sequence have little correlations, a condition that following \cite{tang2020practicality} we call \emph{incoherence}. The idea is that for high-dimensional problems it is natural to expect that gradients sampled in the same batch are likely to have low cosine similarity.

For simplicity, we formulate it pointwise for every sequence $s$ in the population. Assume that there is a bounded $R \geq 1$, such that for all $s \in \mathcal{D}_{tr}$, for all $\xv \in  s$,
\begin{align*}
       \sum_{\xv' \in s} \max_{\hat{y}, \hat{y}'} |\cos(\nabla \ell(\hat{y}, \xv), \nabla \ell(\hat{y}', \xv'))| \leq R\,.
\end{align*}
One way to interpret such condition is as follows: if we look at the cosine similarity as a measure of relevance between the tokens that we predict, we can say that in each sequence there is a limited number of keywords which relate to each other, while the rest of the words are only related locally.

Similar to the independent case, we also assume that for a randomly sampled token, $\nabla \ell(\hat{y}|\xv)$ has a spherical distribution, i.e. $\nabla \ell(\hat{y}|\xv) \overset{d}= \sqrt{N} r H^{1/2} \sv$.

Let $B = s_1 \cup \dots \cup s_b $, where $b$ is the number of sequences in a batch, so that $|B| = b L$, and we assume that each sequence has the same length $L + 1 $ (we do not predict the first token in a sequence, whose index is $0$). Let $\tilde{H}(s)$ denotes in-sequence GNH. Then,
\begin{equation}\label{expansion_batch_to_sequence}
    \E \tilde{H}_t^2 = H^2 + \frac{1}{b} \{ \E \tilde{H}(s)^2 - H^2\},
\end{equation}
where $s$ is a single random sequence. Let us expand,
\[  
    \tilde{H}(s)^2 = \frac{1}{L^2} \sum_{\xv, \xv' \in s} \tilde{H}(\xv) \tilde{H}(\xv')
\]
We have that for $\xv, \xv' \in s$,
\begin{align*}
    \frac{1}{2} &(\tilde{H}(\xv) \tilde{H}(\xv') + \tilde{H}(\xv') \tilde{H}(\xv)) \\
        & = \frac{1}{2} \E_{\hat{y} \sim p(\xv), \hat{y}' \sim p(\xv')} \langle \nabla \ell(\hat{y}|\xv), \nabla \ell(\hat{y}'|\xv') \rangle (\nabla \ell(\hat{y}|\xv)\nabla \ell(\hat{y}'|\xv')^{\T} + \nabla \ell(\hat{y}'|\xv')\nabla \ell(\hat{y}|\xv)^{\T}) \\
        & \preceq
         \frac{\delta(\xv, \xv')}{2}  \left( \E_{\hat{y} \sim p(\xv)} \| \nabla \ell(\hat{y}|\xv)\|^2 \nabla \ell(\hat{y}|\xv)\nabla \ell(\hat{y}|\xv)^{\T} + \E_{\hat{y}' \sim p(\xv')} \|\nabla\ell(\hat{y}'|\xv')\|^{2} \nabla \ell(\hat{y}'|\xv')\nabla \ell(\hat{y}'|\xv')^{\T} \right),
\end{align*}
where we denote for short $\delta(\xv, \xv') = \max_{\hat{y}, \hat{y}'} |\cos(\nabla \ell(\hat{y}, \xv), \nabla \ell(\hat{y}', \xv'))|$, and we also
use the fact that $ 2 \| a \| \| b\| \langle v, a\rangle \langle v, b\rangle \leq \| a\|^{2} \langle v, a\rangle^2 +\| b\|^2 \langle v, b\rangle^2$.
Summing up we have that
\begin{align*}
    \frac{1}{L^2} \sum_{\xv, \xv' \in s} \tilde{H}(\xv) \tilde{H}(\xv') &\preceq \frac{1}{L^2} \sum_{\xv \in s} \left(\sum_{\xv' \in s} \delta(\xv, \xv')\right) \E_{\hat{y} \sim p(\xv)} \| \nabla \ell(\hat{y}|\xv)\|^2 \nabla \ell(\hat{y}|\xv)\nabla \ell(\hat{y}|\xv)^{\T} \\
    & \preceq \frac{R}{L^2} \sum_{\xv \in s} \E_{\hat{y} \sim p(\xv)} \| \nabla \ell(\hat{y}|\xv)\|^2 \nabla \ell(\hat{y}|\xv)\nabla \ell(\hat{y}|\xv)^{\T}
\end{align*}

\begin{figure}[t]
    \centering
    \includegraphics[width=\textwidth]{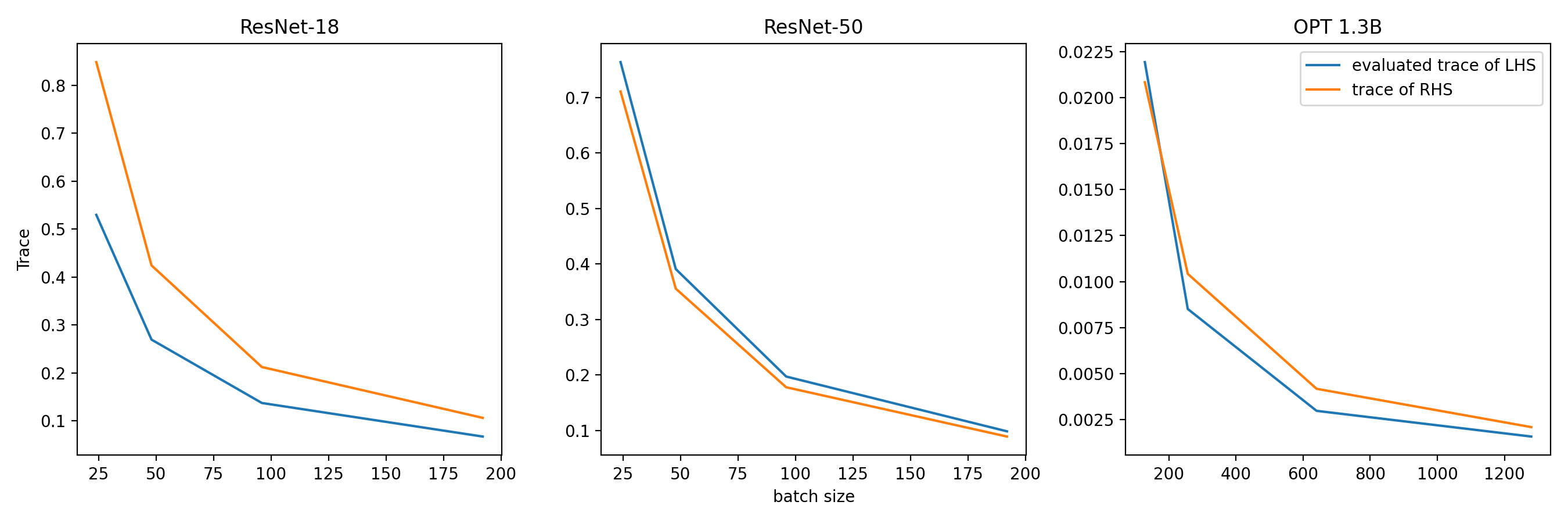}
    \caption{Compring traces of LHS and RHS of condition~\ref{eq:condition1} for different batch sizes. We evaluate the traces for ResNet-18, ResNet-50, and OPT-1.3B, 4 batch sizes for each model. For the OPT-1.3B, the batch size is counted in tokens.}
    \label{fig:condition1}
\end{figure}

From previous section we know that for $\gv = \nabla \ell(\hat{y}|\xv)$, we have $ \E \| \gv\|^{2} \gv \gv^{\T} \preceq (\E r^4) (\Tr(H) H + H^2) $. Taking the expectation over $s \sim \mathcal{D}_{tr}$ and $\xv \sim \mathcal{U}(s)$ we obtain,
\[
    \E \tilde{H}(s)^2 \preceq \frac{R}{L} (\E r^4) (\Tr(H) H + H^2)
\]
Plugging into \eqref{expansion_batch_to_sequence}, we obtain that
\[
    \E \tilde{H}_t^2 - H^{2} \preceq \frac{R}{|B|} (\E r^4) \Tr(H)H + \left(\frac{R}{|B|} (\E r^{4}) - \frac{1}{b}\right) H^2 \,.
\]
\paragraph{Empirical sanity check.}
Furthermore we propose a simple empirical sanity check by assessing the relationship between the LHS and RHS in condition \eqref{eq:condition1}. In Figure~\ref{fig:condition1}, we compare the traces of two matrices, similarly evaluating the traces by averaging over random quadratic forms $ \gv^{\T} (\tilde{H}_t^{2} - H^2) \gv $. We evaluate this gap by estimating the HVPs $\tilde{H}_t\gv$, on random Gaussian vectors and taking their norm. Each evaluation is averaged over 1000 realizations of $\gv$.

\subsection{Counter-example with divergence}\label{counter-example}
\begin{lemma}
There exists a binary classification task, where condition~\ref{eq:condition1} holds with exact equality
\[
    \E \tilde{H}_t^{2} - H^{2} = \frac{1}{|B|} \Tr(H) H,
\]
and for some inputs $\uv^{0}, \gv$ the following claim holds.

Let us choose arbitrary step size satisfying $0 < \eta \leq 1/\lambda_{\max}(H)$, such the matrix 
\[
    (1 - \eta (H + \lambda))^{2} + \eta^{2} \{ \E \tilde{H}_t^{2} - H^{2}\}
\]
has eigenvalue $\lambda \geq 1$. Then, the LiSSA algorithm converges on average, but not samplewise:
\[
    \| \E \uv^{t} - \uv^{\star} \| \rightarrow 0,
    \qquad
    \E \| \uv^{t} - \uv^{\star} \|^{2} \rightarrow \infty \,.
\]

In particular, such $\eta$ exists in the case $ |B| \leq \left(\frac{\lambda_{\max}}{\lambda_{\max} + \lambda}\right)^{2} \frac{\Tr(H)}{\lambda_{\max}}$.
\end{lemma}
\def\sv{\mathbf{s}}
\def\xv{\mathbf{x}}
\begin{proof}
    We consider a binary regression with $y \in \{0, 1\}$ and inputs  $\xv \in \R^{N} $. For simplicity, we assume that $ \xv, y $ are independent, and $y$ takes values $\{0, 1\}$ with equal probabilities. We consider the distribution in $\xv$  that satisfies $ \| \xv \|^{2} = \Tr(H)$ pointwise and $\E \xv\xv^{\T} = H$. 
    For this, let $ H = V \Lambda V^{\T}$ where $ \Lambda = \mathrm{diag}\{ \lambda_1, \dots, \lambda_N\} $  and take $\xv = V \sv$, $\sv = (\sqrt{\lambda_1} \epsilon_1, \dots, \sqrt{\lambda_N} \epsilon_N)^{\T}$ where $\epsilon_i  = \pm1$ with equal probabilities. Consider the binary logistic model
    \begin{align*}
        \log p(y = 1 | \xv) &=\theta^{\T} \xv - \log (\exp(\theta^{\T} \xv) + \exp(- \theta^{\T} \xv)), \\
        \log p(y = 0 | \xv) &= -\theta^{\T} \xv - \log (\exp(\theta^{\T} \xv) + \exp(- \theta^{\T} \xv)),
    \end{align*}
    and assume that during training the model converged to the optimal parameters $ \theta = 0 $, since $\xv, y$ are independent. Then it is straightforward to calculate that
    \begin{align*}
        \tilde{H}_t = \frac{1}{|B|} \sum_{(\xv, y) \in B} \xv \xv^{\T},
        \qquad
        \E \tilde{H}_t^{2} = \left(1 - \frac{1}{|B|}\right) H^2 + \frac{1}{|B|} \Tr(H) H.
    \end{align*}
    Since all matrices are rotated by $V$ from left and right, we assume w.l.o.g. that $V = I$.

    Now assume that $\gv = 0$ we have $\uv^{\star} = 0$, and let $\uv^{0} \neq 0$. We have
    \[
        \uv^{t} = (1 - \eta (\tilde{H}_t + \lambda))  \uv^{t-1} = \prod_{j = 1}^{t} (1 - \eta (\tilde{H}_j + \lambda)) \uv^{0}\,.
    \]
    Set $ Q = I - \eta (\hat{H}_t + \lambda) $ and $ R = \E (1 - \eta (\hat{H}_t + \lambda))^2 = (1 - \eta(H + \lambda))^{2} + \eta^{2} (\E \tilde{H}_t^2 - H^2) $. Notice that both matrices are diagonal. Consider the sequence of matrices $R_0= I$, $R_1 = R$, $ R_k = \E (1 - \eta (\tilde{H}_t + \lambda)) R_{k-1} (1 - \eta (\tilde{H}_t + \lambda)) $. Also denote partial product $\tilde{T}_j = \prod_{k = 1}^{j} (1 - \eta (\tilde{H}_k + \lambda)) $. Then we have
    \begin{align*}
        \E \| \uv^{t} \|^{2} &= [\uv^{0}]^{\T} \E C_{t-1} (1 - \eta (\tilde{H}_t + \lambda))^{2} C_{t-1} \uv^{0} \\
        & = [\uv^{0}]^{\T} \E T_{t-1} R C_{t-1}\uv^{0} \\
        &= [\uv^{0}]^{\T} \E T_{t-2} (1 - \eta (\tilde{H}_{t-1} + \lambda)) R_1 (1 - \eta (\tilde{H}_{t-2} + \lambda)) C_{t-2}\uv^{0}  \\
        &= [\uv^{0}]^{\T} \E T_{t-2}  R_2  T_{t-2}\uv^{0} \\
        &= [\uv^{0}]^{\T} \E T_{t-3}  R_3  T_{t-3}\uv^{0} \\
        &= \dots \\
        &= [\uv^{0}]^{\T} R_t \uv^{0}
    \end{align*}
    Let us show that the  matrix $B_t$ is diagonal. Indeed, assuming $B_{k-1}$ is diagonal, we have that
    \begin{align*}
        R_{k} &= Q R_{k-1} Q + \eta^2 \E (\tilde{H}_t - H) R_{k-1} (\tilde{H}_t - H) \\
        & = Q B_{k - 1} Q - \eta^2 H R_{k-1} H +\eta^2 \E \tilde{H}_t R_{k-1} \tilde{H}_t
    \end{align*}
    We have that for diagonal $R_{k-1}$, $ \xv^{\T} R_{k-1} \xv = \sum_{j} \lambda_j R_{k-1}[j, j] = Tr(HR_{k-1}) $ is deterministic. Therefore,
    \begin{align*}
        \E \left( \frac{1}{|B|} \sum_{x \in B} \xv\xv^{\T} \right) R_{k-1} \left( \frac{1}{|B|} \sum_{x \in B} \xv\xv^{\T} \right)  &= \left(1 - \frac{1}{|B|}\right) H^{2} R_{k-1} + \frac{1}{|B|} \E (\xv^{\T} R_{k-1} \xv) \xv \xv^{\T} \\
        &= \left(1 - \frac{1}{|B|}\right) H^{2} R_{k-1} + \Tr(H R_{k-1}) H  \\
        &= [\E \tilde{H}_{t-1}^{2}] R_{k-1}
    \end{align*}
    Thus, we conclude
    \[
        R_{k} = Q^{2} R_{k-1} - \eta^{2} H^{2} R_{k-1} + \eta^{2} [\E \tilde{H}_t^{2}] R_{k-1} = R R_{k-1} = R^{k} R_{0} = R^{k}.
    \]
    Now, we have that
    \[
        \E \| \uv^{t}\|^{2} = [\uv^{0}]^{\T} R^{t} \uv^{0} \geq \lambda^{t} \langle \uv^{0}, v \rangle^{2} \not\to 0,
    \]
    whenever $\langle \uv^{0}, v \rangle \neq 0$.
\end{proof}

\section{Details of Hessian statistics calculation}\label{sketching_details}

For trace approximation, we use the fact that for a Gaussian vector $ \gv \in \mathcal{N}(0, \tfrac{1}{N} I) $, generated independently from $\tilde{H}$, \(\E \gv^{\T}\tilde{H} \gv = \frac{1}{N} \Tr(H) \,\).
Observe that from~\eqref{formula-tilde-H-dot-u},
\[  
    \gv^{\T}\tilde{H} \gv = \sum_{j = 1}^{N} \left\{ J_{\theta} h(\xv; \theta)^{\T} \gv \right\}^{\T} \{ \mathrm{Diag}(\sf(h)) - \sf(h)\sf(h)^{\T} \} [ J_{\theta} h(\xv; \theta)^{\T} \gv ],
\]
thus evaluating such quadratic form can be done by only evaluating the middle Hessian $ \mathrm{Diag}(\sf(h)) - \sf(h)\sf(h)^{\T} $ and finite differences $J_{\theta} h(\xv; \theta)^{\T} \gv \approx 50(h(\xv; \theta + 0.01 \gv) - h(\xv; \theta - 0.01 \gv)) $. The results reported in Table~\ref{hessian-stats-table} are based on 1600 evaluations of $ \gv_{i}^{\T} \tilde{H}_i \gv_i $, including the standard error.

At an increased price, we can also evaluate the Frobenius norm of the Hessian. For that, for a given vector $\gv$, sampled from Gaussian distribution, we evaluate two independent Hessian-vector products $ \tilde{H} \gv $, $\hat{H}\gv$, so that $ \frac{1}{N} \Tr(H^2) = \E (\tilde{H} \gv)^{\T} \hat{H}\gv  $. Then, by sampling independently a series of independent realizations, we can evaluate the mean $\frac{1}{N} \Tr(H^2)$ and the standard error of our estimation. We report these evaluations in Table~\ref{hessian-stats-table-second}.

For evaluating the top eigenvalues, one can employ the sketching technique. For example, \cite{swartworth2023optimal} show that for $\Phi \in \R^{d \times N}$ generated in a way such that $ \Phi_{ij} \sim \mathcal{N}(0, \tfrac{1}{d}) $ for $d$ large enough, we have that for finite amount of top eigenvalues,
\begin{equation}\label{swartworth}
    \lambda_{l}(H) \approx \lambda_{l}(\Phi H \Phi^{\T}).
\end{equation}
In order to evaluate the matrix $ \Phi H \Phi^{\T}$, we iterate over each of $d$ columns ${\boldsymbol \phi}_j$ of $\Phi^{\T}$ and evaluate the HVP $H {\boldsymbol \phi}$ by sampling empirical $ \tilde{H}_t {\boldsymbol \phi} $. For language models, to speed up the caluculation we truncate the context length to $256$ and average over batch of size $50$. For image classification, we use a batch size of $ 5000$ to evaluate each $H {\boldsymbol \phi}$. We then project each of these columns back with the embedding $\Phi$, so the result is a $d \times d $ matrix, whose top eigenvalue can be calculated with standard linear algebra packages. According to \cite{swartworth2023optimal},
the error term in \eqref{swartworth} is bounded by multiple of $ \| H \|_{Fr}  / \sqrt{d} $. Although we do not offer a precise control, we suggest that taking $d = 5000$ should be sufficient for the models in Table \ref{hessian-stats-table-second}. Note that their bound does not account for sampling error and we leave it out of consideration in the scope of this paper, we simply want to produce some adequate bound on the largest eigenvalue.
\def\gv{\mathbf{g}}
\def\Phiv{{\boldsymbol{\{\Phi\}}}}

\begin{table}[t]
\centering
\begin{threeparttable}
  \caption{Second moment statistics for  Gauss-Newton-Hessian calculated on ImageNet (IN) on vision and Open-Web-Text-2 (OWT) on language models.}
  \label{hessian-stats-table-second}
  \begin{tabular}{lrrrrl}
    \toprule
    Model     & Size & Data&  $\| H \|_{Fr}$     & $\lambda_{\max}(H)$ & \makecell{$\left(\frac{\| H \|_{Fr}}{\lambda_{\max}(H)}\right)^{2}$} \\
    \midrule
    ResNet-18 & 11M & IN & $2.55 \times 10^{4}$ & $\approx 270$ & $8.92 \times 10^{3}$   \\
    ResNet-50 & 25M & IN  & $2.67 \times 10^{4}$ &  $\approx 470$ & $3.22 \times 10^{3}$  \\
    OPT &1.3B & OWT    & $2.94 \times 10^{3}$  & $\approx 780$  & $1.42 \times 10^{1}$  \\
    Llama-1 & 7B  & OWT   &   $3.73 \times 10^{3}$     & $\approx 1600$ & $5.43$ \\
    Mistral & 7B & OWT    &   $2.49\times 10^{4}$    & $\approx 5600$ & $1.98 \times 10^{1}$  \\
    \bottomrule
  \end{tabular}
  \end{threeparttable}
\end{table}

\paragraph{Pseudo-random embeddings.} Generating and storing a dense embedding matrix $ \Phi \in \R^{d \times N} $ can be prohibitively expensive, since each of the rows $\Phi[i, :]$ is equivalent to one model in memory. We propose instead to  use pseudo-random generators, and generate $\Phi$ ``on-the-fly'' using a single integer number \emph{seed}. The model parameters are usually accessed in a form of lists $ \theta^{\T} = (\theta_1^{\T}, \dots, \theta_L^{\T}) $ say where $L$ is the number of layers. Correspondingly, the gradients and HVPs are also iterated over a list $ \gv = (\gv_1, \dots, \gv_L)$. We generate $\Phi$ in the form $\Phi = \begin{pmatrix} \Phi_1 & \dots & \Phi_L \end{pmatrix}$, so that $ \Phi \gv = \sum_{j = 1}^{L} \Phi_j \gv_j$. To calculate $\Phi \gv $ we initialize a random generator with a fixed seed, and then generate $ \Phi_j$ on the fly, so that we never have to store the whole matrix $\Phi$ in the memory.

Nevertheless, the embedding operation itself $\Phi\gv$ can be too expensive. Our observation is that starting $d = 50$ we do not really benefit from parallel matrix computations on GPU and the price scales linearly, i.e. $d=200$ is $4$ times as expensive as $d = 50$, and so on. Furthermore, the price of a single application of embedding for $d = 50$ can be as high as the gradient computation itself. In order to reduce the computational price of embedding, we suggest to use the following heuristic. 
Instead of summing up per-layer embeddings
\(
    \Phi \gv = \Phi_1 \gv_1 + \dots + \Phi_{L} \gv_{L}
\), we suggest to concatenate them.
This way, we can increase the  dimension by a factor of $L$ with little additional computation. Let us denote the resulting concatenating embedding by $\Phiv$, then we can write
\begin{align*}
    \Phi &= \begin{pmatrix}
        \Phi_1 & \Phi_2 & \dots & \Phi_L
    \end{pmatrix} \in \R^{d \times N} \\
    \Phiv &= \begin{pmatrix}
        \Phi_1 & 0 & \dots & \\
        0 & \Phi_1 & \\
         & & \dots &\\
         0 & 0 & \dots & \Phi_{L}
    \end{pmatrix}
    \in \R^{Ld \times N}
\end{align*}
In other words, we have $ \Phiv \theta = \mathtt{vstack}([\Phi_1\theta_1, \dots, \Phi_L \theta_L])$. That is, each $\Phi_1 \theta_1$ has dimension $d$ and
\[
    \E_{\Phi} [\Phiv \theta]^{\T}[\Phiv \theta'] = \sum_j \E [\Phi_j \theta_j]^{\T} [\Phi_j \theta_j'] = \sum_j \theta_j^{\T} \theta_j' = \theta^{\T} \theta',
\]
so that on average it preserves the dot products as well. However, concatenation can dramatically reduce the variance of one dot product. Indeed, we have that
\begin{align*}
    \Var_{\Phi}([\Phiv \theta]^{\T}[\Phiv \theta']) &= \sum_{j} \Var( [\Phi_j \theta_j]^{\T} [\Phi_j \theta_j']) \lesssim \frac{1}{d} \sum_{j}  \| \theta_j\|^{2} \| \theta_j'\|^2,
\end{align*}
and recall the bound from before,
\[
    \Var_{\Phi}([\Phi \theta]^{\T}[\Phi \theta'])  \lesssim \frac{1}{d}  \| \theta\|^{2} \| \theta'\|^2 = \frac{1}{d} \left( \sum_{j} \| \theta_j\|^{2} \right) \left(\sum_{j} \| \theta_j'\|^{2} \right)
\]
The former can be much smaller than the latter when square norms of the gradients are spread ``evenly'' over the layers. That is, assume that $\| \theta_j\|^2$ is approximately in the same bulk $ C^{-1} M \leq \| \theta_j\|^{2} \leq C M $, $M = \frac{1}{L} \sum_{j} \| \theta_j\|^{2}$. Then $ \Var_{\Phi}([\Phiv \theta]^{\T}[\Phiv \theta']) \lesssim \frac{L}{d} M^2 $ while $ \Var_{\Phi}([\Phi \theta]^{\T}[\Phi \theta']) \lesssim \frac{L^2}{d} M^2 $, so we effectively it is equivalent to increasing dimension $L$ times with original Gaussian features.
Here we ignored the terms $\langle \theta, \theta'\rangle^2$ but in practice they are significantly smaller than $\| \theta\|^{2} \|\theta'\|^{2}$.

%We can further increase the dimension 4 times if we concatenate according to how each tensor split in the multi-gpu setting. Say if we use 4-GPU tensor splitting we can have 4x dimension for free, since these computations are done separately anyway.

\section{Influence functions for bag-of-words model is TF-IDF}\label{section-tf-idf}
\def\sf{\mathrm{sf}}
\def\Diag{\mathrm{Diag}}
\def\pv{\mathbf{p}}
\def\qv{\mathbf{q}}

Recall that for a set of documents $d \in D$ and terms $t \in T$. Define term frequency (TF) and document frequency (DF) as follows,
\begin{align*}
    TF(t, d) &= \frac{1}{|d|} count(t, d), \\
    DF(t) &= \frac{\# \left\{ d \in D: t \in d \right\}}{|D|};
\end{align*}
let us also consider a variant of inverse document frequency $ IDF(t) = \sqrt{1/DF(t)} $. Note that in standard $TF{\cdot} IDF$ definition, the square root is replaced with logarithm, we only propose to consider the square root for sake of comparison to influence functions. Then, the document-term relevance $TF\cdot IDF$ is calculated as the product of $TF$ and $IDF$, and the corresponding similarity between $d_1$ and $d_2$ is
\[
    \mathrm{sim}(d_1, d_2) = \sum_{t} TF\cdot IDF(d_1, t) TF\cdot IDF(d_2, t) = \sum_{t} \frac{1}{DF(t)} TF(d_1, t) TF(d_2, t) 
\]

The bag-of-words model reads as follows,
\[
    \log p(d) = \sum_{t} \mathtt{count}(t, d) \log p_t,
\]
where $p_t = \exp(x_t) / (\exp(x_1) + \dots + \exp(x_T))$, the $x_t$ are parameters. For sake of simplicity we assume that each document has the same length $|d|$.  We have,
\begin{align}
    \nabla_{x} \log p(d) &=  \sum_{t} {count}(t, d) e_t - |d| \nabla_{x} \log \left( \sum_t \exp(x_t)\right) = \sum_{t} {count}(t, d) e_t  - |d| \sf(x) \nonumber \\
    &= |d| \left\{ \sum_{t} TF(t, 
    d) e_t - \sf(x) \right\} \label{tf-idf-ad-hoc}
\end{align}
Notice that by definition, $\mathbf{1}^{\T} \nabla_{x} \log p(d)= 0$.
The Hessian looks as follows
\[
    \nabla_{x}^{2} \E \log p(d) = - |d| \nabla_{x}^{\T} \sf(x) = - |d| \left( \Diag(\sf(x)) - \sf(x) \sf(x)^{\top} \right),
\]
where we calculate that
\[
    \nabla_{x_i} \frac{\exp(x_j)}{\sum_k \exp(x_k)} = \frac{\delta_{ij}\exp(x_i)(\sum_k \exp(x_k)) - \exp(x_i)\exp(x_j) }{(\sum_k \exp(x_k))^{2}} = \delta_{ij} \sf(x)_i - \sf(x)_i \sf(x)_j
\]
\def\rv{\mathbf{r}}
Let us calculate the inverse Hessian. Given a damping parameter $\lambda$, let $\qv = (\pv + \lambda)^{1/2}$ elementwise. Note that $\| \qv \|^{2} = 1 + \lambda N$. Set also $ \rv = \pv / (\pv + \lambda)^{1/2}$ elementwise, and notice that $\| r\|^{2} = \sum_{j} p_j^2 / (p_j + \lambda) = 1 - \lambda \sum_{j} p_j / (p_j + \lambda)  < 1 $. Then,
\[
    \Diag(\pv) + \lambda I - \pv \pv^{\T} = \Diag(\qv) (I - \rv \rv^{\T}) \Diag(\qv),
\]
so the inverse equals to
\def\dv{\mathbf{d}}
\begin{align*}
    (H + \lambda)^{-1} &= \Diag(\qv)^{-1} \left(I + \frac{1}{1 - \| \rv\|^{2}} \rv \rv^{\T} \right) \Diag(\qv)^{-1} \\
    &= \Diag(\pv + \lambda)^{-1} - \left(\lambda \sum_j p_j / (p_j + \lambda)\right)^{-1} \dv \dv^{\T},
\end{align*}
where we denote $\dv = \pv / (\pv + \lambda) = \mathbf{1} - \lambda / (\pv + \lambda)$ elementwise.
Note that the gradients $\gv_d = \nabla \log p(d)$ are in the subset $\gv_d^{\T} \mathbf{1}$. Since $ \dv = \mathbf{1} + O(\lambda)$ we have that
\[
    \gv_{d_1}^{\T}\left(\lambda \sum_j p_j / (p_j + \lambda)\right)^{-1} \dv \dv^{\T} \gv_{d_2} = O(\lambda),
\]
therefore, in the limit $\lambda \rightarrow 0$, we have that the influence between documents $d_1, d_2$ reads as
\[
    \mathcal{I}(d_1, d_2) = \gv_{d_1}^{\T} \Diag(\pv)^{-1} \gv_{d_2} = \sum_{t} TF(t, d_1) TF(t, d_2) p_t^{-1}\,.
\]

Let us also calculate the IDF for this models. Let us assume that terms are rare enough ( $p_t |d| \ll 1$). Then we have that,
\begin{align*}
    DF(d) &= 1 - \Pr(t \notin d) = 1 - (1 - p_t)^{|d|} \approx |d| p_t\,.
\end{align*}
and therefore up to a scaling factor, the above expression is approximately equal to the $TF\cdot IDF$ relevance in \eqref{tf-idf-ad-hoc}.

\begin{comment}
\section{Nesterov acceleration}
\def\mv{\mathbf{m}}
\def\vv{{\mathbf{v}}}
\def\pv{{\mathbf{p}}}

Set
\begin{align*}
    \mv^{1} &= 0 \\
    \pv^{t} &= (\tilde{H}^{t} + \lambda)\uv^{t-1} - \gv \\
    \mv^{t} &= \alpha \mv^{t-1} + \pv^{t} \\
    \uv^{t} &= \uv^{t-1} + \eta \mv^{t}
\end{align*}
Write
\begin{align*}
    \uv^{t} - \uv^{\star} &= \eta \alpha \mv^{t-1} + \eta\alpha \vv^{t} \\
    &= (1 - \alpha) (\uv^{t-1} - \uv^{t-2}) + \alpha  \eta \left( (1 - \lambda \eta) I + \eta \tilde{H}_t\right) \uv^{t-1} + \alpha \eta \gv \\
    &= (1 - \alpha) (\uv^{t-1} - \uv^{\star}) + \alpha \left( I - \eta (\tilde{H}_t + \lambda)\right) (\uv^{t-1} - \uv^{\star}) + \alpha \eta (H - \tilde{H}_t) \uv^{\star} \\
    &=  \left( I - \alpha \eta (\tilde{H}_t + \lambda)\right) (\uv^{t-1} - \uv^{\star}) + \alpha \eta (H - \tilde{H}_t) \uv^{\star}
\end{align*}
\textcolor{red}{Apparently NAG does not work well with small batch size. Makes sense!}
\end{comment}

\section{More examples for comparison of LiSSA and PBRF}\label{sec:appendix_pbrf}

\begin{figure}[t]
    \centering
    \includegraphics[width=\textwidth]{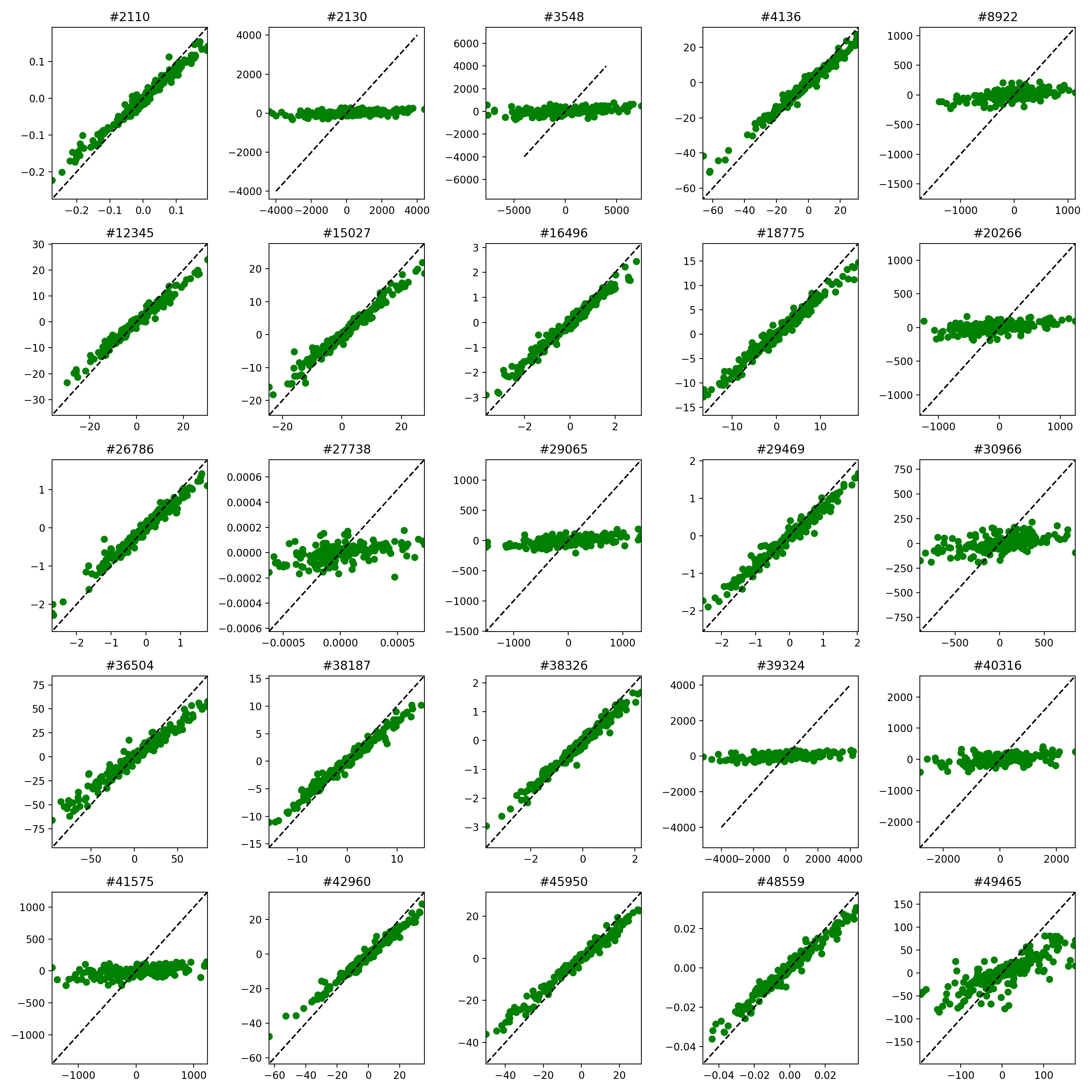}
    \caption{Comparison of PBRF and LiSSA influence on ResNet-18 for 25 random train images. Each graph shows influence of one train image w.r.t. to 500 other test images. Reference number is show above the image, refer to Figure~\ref{fig:retrain_examples}. The results are for ResNet-18, the $x$-axis is LiSSA, and the $y$-axis is PBRF.
    }
    \label{fig:retrain_rn18}
\end{figure}

\begin{figure}[t]
    \centering
    \includegraphics[width=\textwidth]{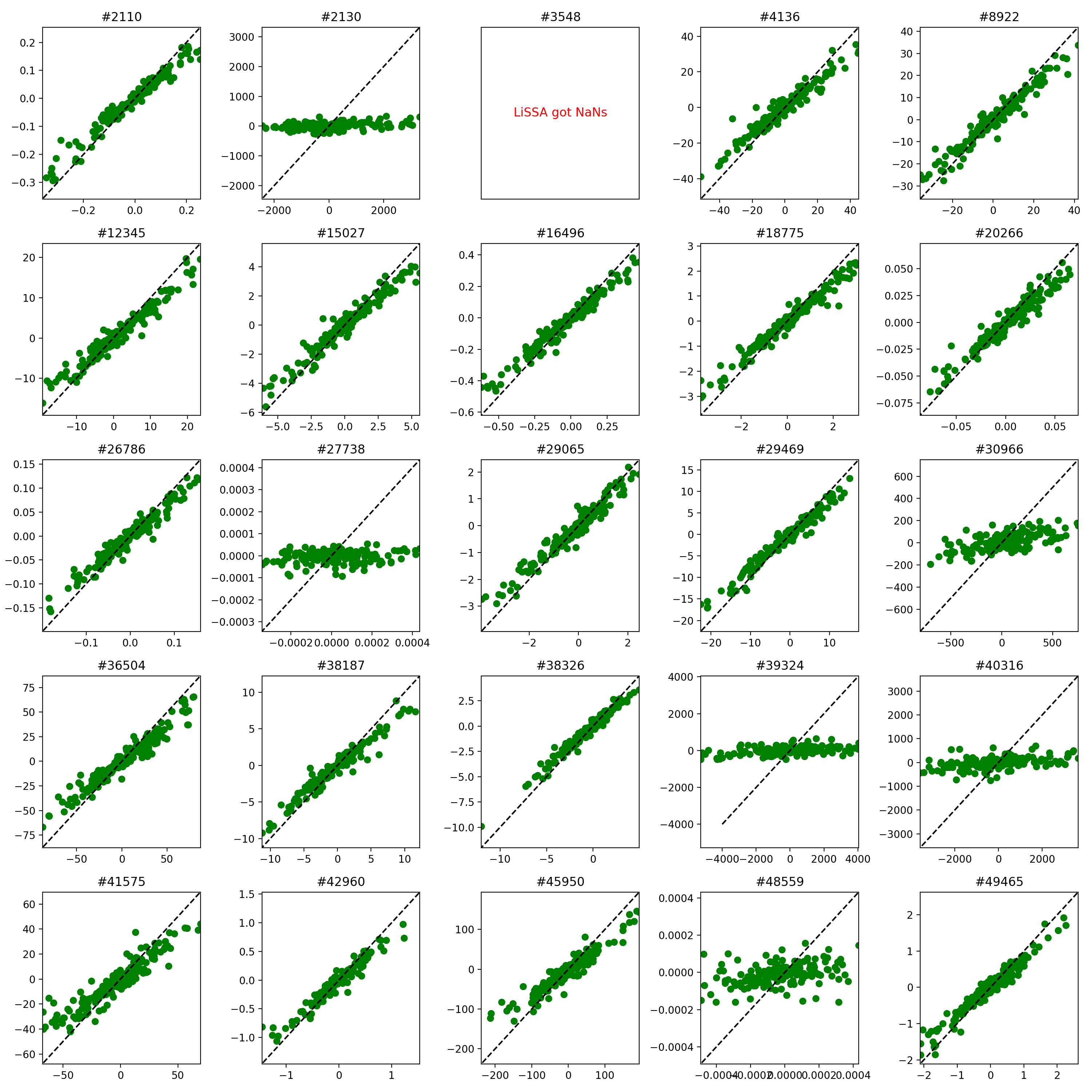}
    \caption{Same as Figure~\ref{fig:retrain_rn18}, but for ResNet-50.
    }
    \label{fig:retrain_rn50}
\end{figure}

\begin{figure}[t]
    \centering
    \includegraphics[width=\textwidth]{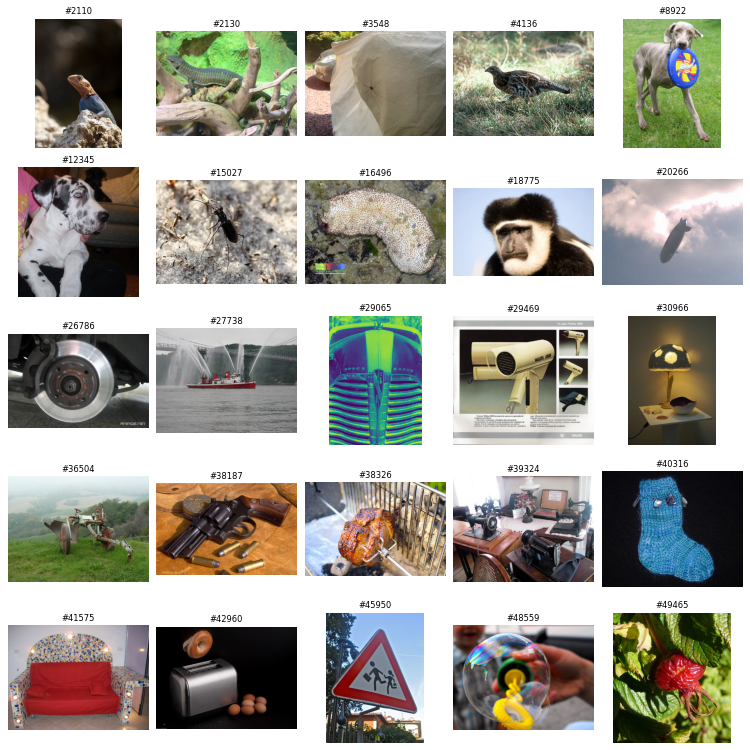}
    \caption{Reference images from ImageNet. Numbers above each image corresponde to those Figures~\ref{fig:retrain_rn18}, \ref{fig:retrain_rn50}.
    }
    \label{fig:retrain_examples}
\end{figure}

Here we present a complete list of 25 train images for comparison of PBRF and LiSSA. We calculate the LiSSA according to the hyperparameter recommendations in Table~\ref{hessian-stats-table}. Figure~\ref{fig:retrain_rn18} shows scatter plots of LiSSA and PBRF for ResNet-18, and Figure~\ref{fig:retrain_rn50} shows scatter plots for ResNet-50. Figure~\ref{fig:retrain_examples} shows reference images from the ImageNet dataset. Note that for one train image in Figure~\ref{fig:retrain_rn50} the LiSSA got float overflow, we attribute it to the high value of the gradient norm.

\end{document}